\pdfoutput=1

\documentclass[11pt]{article}

\usepackage[table]{xcolor} 

\usepackage[]{ACL2023}

\usepackage{times}
\usepackage{latexsym}

\usepackage[T1]{fontenc}

\usepackage[utf8]{inputenc}

\usepackage{microtype}

\usepackage{inconsolata}

\usepackage{amsmath}
\usepackage{amssymb}
\usepackage{booktabs}
\usepackage{multirow}
\usepackage{graphicx}
\usepackage[ruled,vlined]{algorithm2e}
\usepackage{algorithmic}
\usepackage{color}
\usepackage{graphicx}
\usepackage{etoolbox}
\usepackage{spverbatim}
\usepackage{booktabs} 
\usepackage{bbding}
\usepackage{threeparttable}
\usepackage{longtable}
\AtBeginEnvironment{spverbatim}{\small}

\usepackage{times}
\usepackage{latexsym}
\usepackage{amsmath}
\usepackage{makecell}
\usepackage{tabularx}
\usepackage{tablefootnote}

\newcommand{\modelname}{\textsc{Erabal}\xspace}

\usepackage{colortbl} 

\definecolor{LightGold}{RGB}{246,190,100}
\definecolor{LimeGreen}{RGB}{162,248,98}
\definecolor{LightRed}{RGB}{246,100,107}

\title{ERABAL: Enhancing Role-Playing Agents through \\Boundary-Aware Learning}

\author{ 
Yihong Tang\textsuperscript{\rm 1}\thanks{\ \ Equal contribution.}, Che Liu\textsuperscript{\rm 1}$^*$, Jiao Ou\textsuperscript{\rm 1},  Fuzheng Zhang\textsuperscript{\rm 1}, \\ {\bf Di Zhang\textsuperscript{\rm 1}, Kun Gai\textsuperscript{\rm 1}}\\ 
\textsuperscript{\rm 1} Kuaishou\\
{ \{tangyihong,liuche03\}@kuaishou.com}
}

\begin{document}
\maketitle

\begin{abstract}
Role-playing is an emerging application in the field of Human-Computer Interaction (HCI), primarily achieved through the alignment training of large language models (LLMs) with various characters.
Despite significant progress, role-playing agents (RPLAs) still encounter challenges in maintaining character trait and factual consistency during conversations, particularly when confronted with boundary queries subtly related to character attributes.
In this paper, we present \modelname, a boundary-aware learning framework designed to enhance the role-playing capabilities of RPLAs.
\modelname encompasses a data generation pipeline and a concomitant alignment training methodology, aimed at generating and learning from boundary samples respectively.
Through comprehensive evaluations, \modelname demonstrates significant improvements across WikiRoleEval, CharacterEval, and the role-playing subset of MT-Bench, outperforming generalist baseline models while requiring considerably fewer dialogues.
Our code and datasets will be made publicly available to support further research.
\end{abstract}

\section{Introduction}
\begin{figure}[htbp]
\centering
\resizebox{.48\textwidth}{!}{
\includegraphics[width=.48\textwidth]{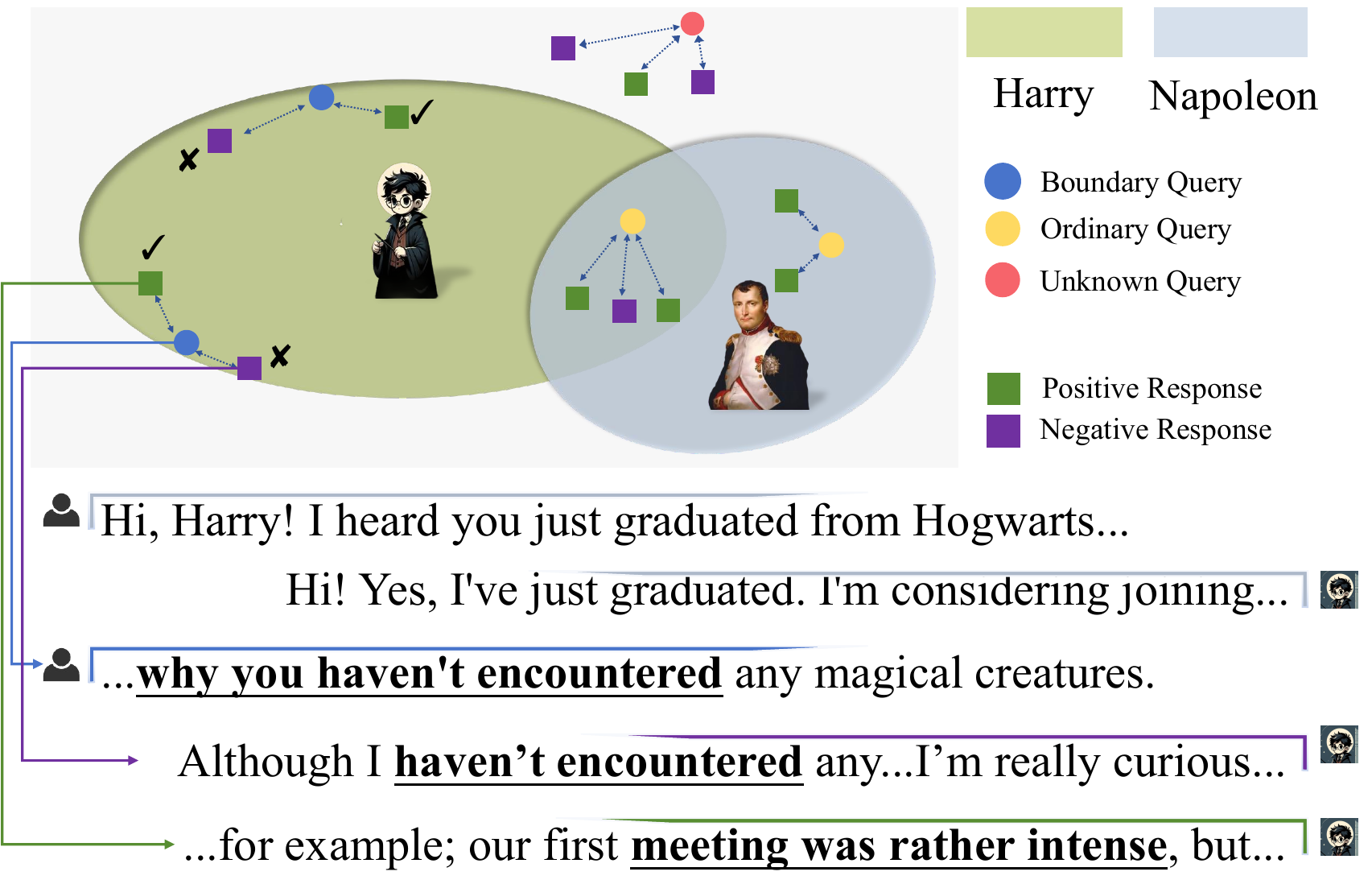}}
\caption{
Illustration of boundary-aware learning.
The circular areas represent the cognitive scope of the characters.
High-value queries (marked in blue) are located within and near the boundaries, while ordinary queries (marked in yellow) are closer to the center.
Unknown queries (marked in red), which should be dismissed with rejection responses, fall outside the circles.
Both positive and negative responses are generated in our work.
}
\label{fig:ex}
\vspace{-0.5cm}
\end{figure}

Role-playing is an emerging research field~\cite{Chen2024FromPT, Tseng2024TwoTO} where AI agents engage in dialogues with humans by simulating the linguistic styles or behavioral traits of the assigned characters~\citep{Ouyang2022TrainingLM, Yi2024ASO}.
As large language models (LLMs) have shown great potential for achieving human-like intelligence, their use in role-playing agents (RPLAs)~\citep{Shanahan2023RolePW, Mao2023EditingPF, Zhou2023CharacterGLMCC, Li2023ChatHaruhiRA, Park2023GenerativeAI} has garnered increasing attention, leading to remarkable progress recently.
Nonetheless, RPLAs still struggle to maintain consistency during human-computer dialogues, facing issues such as character trait contradictions and factual errors.
Figure~\ref{fig:ex} shows such a case, where the RPLA takes on the character of Harry Potter and confronts with the query about ``magical creatures".
Models that have not been enhanced by boundary-aware learning are prone to falling into traps, resulting in out-of-character (OOC) issues.

There have been various efforts towards addressing this challenge~\citep{shao-etal-2023-character, Yu2024NeekoLD, Lu2024LargeLM}.
A notable approach among them, proposed by Ditto\citep{Lu2024LargeLM}, constructs training dialogues with inter-role contrastive queries.
These queries are devised to be answerable for specific roles but fall outside the cognitive scopes of others, as they transcend their era or occupation.
By drawing rough boundaries for characters through alignment training, Ditto demonstrates its effectiveness in improving RPLAs' role-playing capabilities.
However, since this method is relatively coarse-grained, there is still room for improvement by training through a more fine-grained strategy.


In this paper, we propose \modelname, a boundary-aware learning framework aimed at enhancing RPLAs’ role-playing capabilities.
\modelname introduces an automatic data generation pipeline consisting of four modules: a dialogue planner, a topic manager, a query generator, and a response generator.
During training data construction, the query generator first selects a snippet from a comprehensive external character attribute resource (e.g., Wikipedia\footnote{\url{www.wikipedia.org}}) and then tampers it as a counterfactual prerequisite to prompt an LLM (e.g., GPT-4) to generate a role-specific query.
The character attribute resource is not visible to LLMs during training, and a short role description is provided instead.
Such queries are boundary-aware since they are highly role-specific and cannot be easily answered through the general knowledge in LLMs.
The original snippet is further passed to the response generator, and is leveraged to generate a factual response that is consistent with the assigned character.
Dialogues consisting of such boundary-aware query-response pairs are used as supervised fine-tuning data.
To depict fine-grained boundaries for roles, we also introduce a post-training stage.
We pair each aforementioned factual response with a counterfactual response as the negative sample by providing the response generator with the tampered snippet.
The pairs, along with their dialogue context, are utilized by the direct preference optimization (DPO)~\citep{Rafailov2023DirectPO} method.

\begin{table}[t!]
\centering
\resizebox{.47\textwidth}{!}{
\begin{tabular}{l|cc}
\toprule  

       \makecell[c]{Models} & \makecell[c]{Boundary \\ Role Consistency} & \makecell[c]{MT-Bench \\ Role-play} \\ 
        \midrule
        Baichuan2-13B & 0.44 & 7.60 \\
        \quad\textit{+}RoleBench & 0.42(-0.02) & 7.80(+0.20)  \\
        \quad\textit{+}Ditto & 0.48(+0.02) & 7.80(+0.20) \\
        \quad\textit{+}\modelname & \textbf{0.62(+0.18)} & \textbf{8.05(+0.45)}  \\
        \midrule
        Mistral-7B & 0.45 & 8.20 \\
        \quad\textit{+}RoleBench & 0.46(+0.01) & 8.30(+0.10)  \\
        \quad\textit{+}Ditto & 0.52(+0.07) & 8.35(+0.15) \\
        \quad\textit{+}\modelname & \textbf{0.61(+0.16)} & \textbf{8.50(+0.30)}  \\        
\bottomrule
\end{tabular}}
\caption{
Experimental results of Baichuan2-13B-Chat and Mistral-7B-Instruct-v0.3 in boundary scenario evaluation (Section~\ref{sec: bse}) and MT-Bench role-playing evaluation.
\modelname achieves an average of 0.17 and 0.38 absolute improvements over the generalist baselines.
}\label{tab: intro_case}
\vspace{-0.5cm}
\end{table}

To serve as a proof of concept, we first collect 90 popular roles and 500 boundary-aware dialogues as the test set using \modelname.
We evaluate generalist models and their variants through the role consistency evaluation proposed in Ditto.
As shown in Table \ref{tab: intro_case}, generalist models such as Baichuan2-Chat~\cite{Yang2023Baichuan2O} and Mistral-Instruct~\cite{jiang2023mistral} perform sub-optimally, while those enhanced with contrastive queries indeed demonstrate superior performance.
In comparison, variants enhanced with \modelname demonstrate the best performance (will be detailed in Section \ref{sec: bse}).
We also conduct comprehensive experiments across WikiRoleEval~\citep{Lu2024LargeLM}, CharacterEval~\citep{Tu2024CharacterEvalAC}, and the role-playing subset of MT-Bench \citep{Zheng2023JudgingLW}.
In this case, we additionally collect a training dataset containing 190 roles and 16,932 dialogues, which includes 100 roles that are identical to those in RoleBench\footnote{RoleBench has only 5 Chinese roles; we collect an additional 90 Chinese roles for the comprehensive evaluation.} \citep{Wang2023RoleLLMBE}.
Experimental results demonstrate that \modelname is both efficient and effective: by using only 10\% of the training dialogues compared to those in RoleBench (~168,093), \modelname achieves significant improvements over the generalist baselines across the three benchmarks, highlighting a clear advantage of our method over previous approaches.

Our contributions are as follows: (1) We introduce the boundary-aware learning method and demonstrate its effectiveness in developing role-playing agents. (2) We propose \modelname, a framework encompassing a multi-module data production pipeline and the concomitant methodology for supervised fine-tuning and post-training. (3) Extensive experiments show that our method significantly enhances performance across existing role-playing benchmarks, particularly enhancing metrics closely related to role-playing abilities.

\section{Boundary-aware Dialogue Construction}
\begin{figure*}[htbp]
\centering
\includegraphics[width=1\textwidth,keepaspectratio]{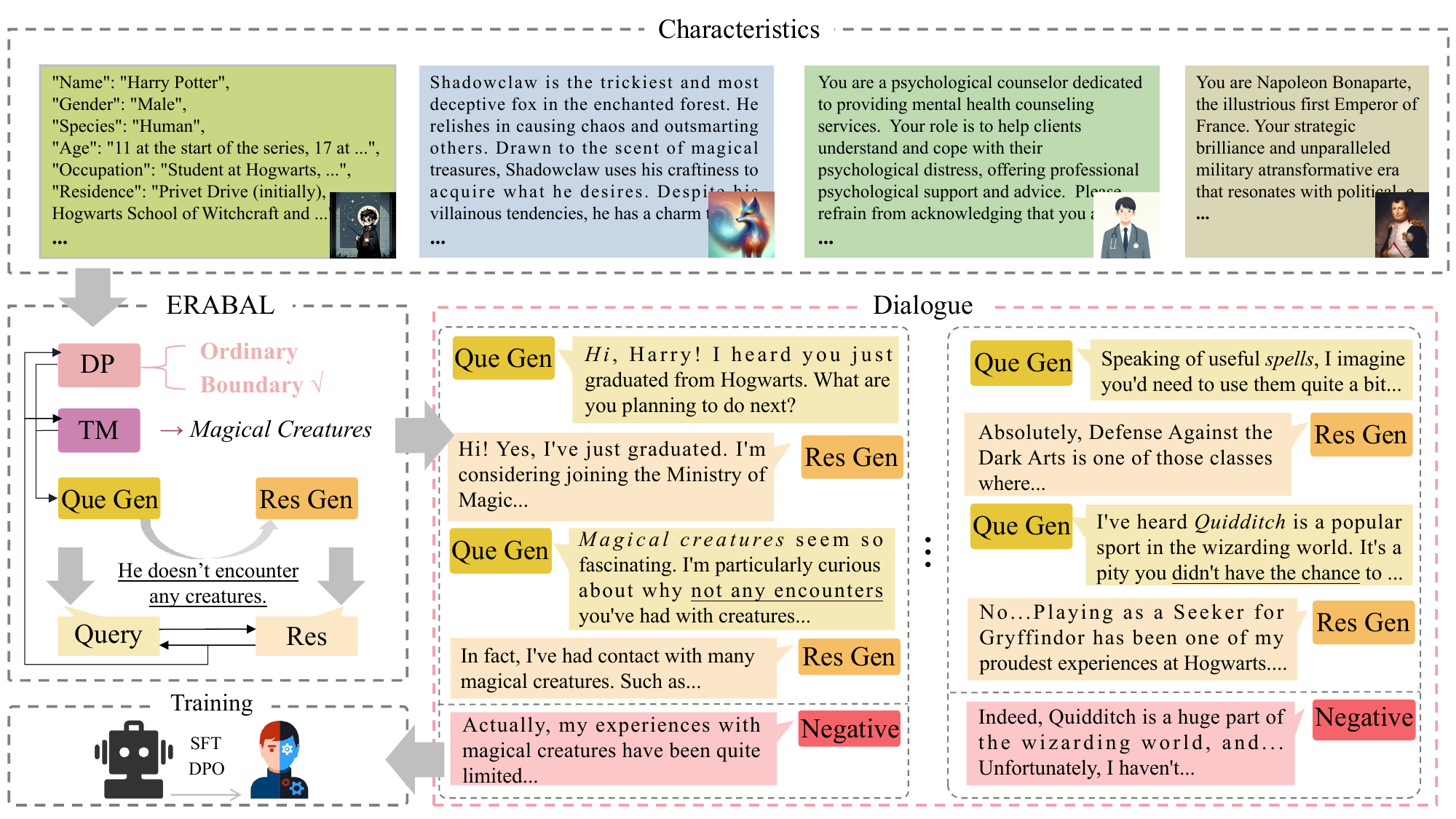}
\caption{
The overview architecture of \modelname.
DP, TM, Que Gen, and Res Gen refers to the dialogue planner, the topic manager, the question generator, and the response generator.
The topic related words are presented in italics, and the counterfactual information in boundary-aware questions is underscored.
The factual (positive) and counterfactual (negative) responses are colored \textcolor{LightGold}{lightgold} and \textcolor{LightRed}{red}, respectively.}

\label{fig:overview}
\vspace{-0.5cm}
\end{figure*}

Figure~\ref{fig:overview} shows the overall data generation process.
To construct a boundary-aware dialogue, we begin with the dialogue planner and topic manager.
These two modules generate instructions at the start of each turn, determining the question type and specifying the conversational topic.
The question type can be either ``ordinary'' or ``boundary''.
Ordinary questions are introduced to ensure that the constructed dialogue remains coherent, which is essential for maintaining contextual relevance during conversations.
The instructions are then passed to the query generator and the response generator, which are responsible for generating specific question-answer pairs based on external character attributes.
The external character attributes are replaced with a brief description in the final training data.
In the following section, we will first describe each component, then introduce the quality control strategies, and finally provide an overall statistic of the constructed dataset.

\subsection{Main Components}
\paragraph{Dialogue Planner}
The dialogue planner, motivated by~\citep{Yuan2023DistillingSK}, functions as a binary switch that provides an instruction for the dialogue state at the beginning of each turn.
The instruction it outputs is either ``ordinary'' or ``boundary'', determined by the given character attributes, dialogue context, and historical planning states.
The ordinary output instructs the query generator to produce a general role-specific query, such as chitchat or banter, while the boundary output guides the construction of the query with a potential trap (i.e., the boundary query).
We prompt the dialogue planner to generate boundary outputs as frequently as possible while allowing them to navigate between ordinary and boundary states to maximally maintain the dialogue coherency.
In practice, we observe about 65.44\% of all dialogue turns are instructed with boundary.
The detailed implementation is shown in Table~\ref{tab: dia_plan}.

\vspace{-0.1cm}

\paragraph{Topic Manager}
The topic manager plays a crucial role in maintaining the relevance of the conversational content to character attributes throughout the dialogue.
Taking the identical inputs as the dialogue planner, it dynamically adjusts the topics at the beginning of each turn, and decide whether to deepen or expand the topic as the dialogue progresses.
It is also prompted to ensure that the generated dialogues are diverse, referring to as many topics in the character's attributes as possible.
More details are shown in Table~\ref{tab: top_man}.

\paragraph{Query Generator}
The query generator takes character attributes and dialogue context as inputs and simultaneously accepts the results derived from the dialogue planner and the topic manager.
It first selects a snippet from the character attributes according to current topic and tempers it if necessary according to the instructional output of the dialogue planner.
Then, the tempered snippet acts as a prerequisite, compelling the query generator to produce a query that conceals a piece of counterfactual information.
Regardless of whether tempering actually occurs, a query is eventually generated and sent to the response generator.
Detailed prompts for the snippet selection and query generation process are shown in Table~\ref{tab: conter_ope} and Table~\ref{rab: que_gen}.

\vspace{-0.1cm}

\paragraph{Response Generator}
The response generator is responsible for generating responses based on the given character attributes, conversational context, and snippets.
Note that snippets are passed to this module as references only when the boundary instruction are specified.
The response generator is expected to produce both factual and counterfactual responses corresponding to the original and tampered snippets respectively.
In practice, even if we provide an original snippet, the response generator may still produce a counterfactual response, and vice versa. 
Therefore, we additionally introduce an information verifier to perform secondary verification to ensure the responses meet our demands.
Detailed prompts are listed in Table~\ref{tab: res_gen} and Table~\ref{tab: info_veri}.


\begin{table}[t!]
\centering
\small
\resizebox{.49\textwidth}{!}{
\begin{tabular}{lc}
\toprule
\rowcolor{white} \textbf{Quality Review Questions for Human} & \textbf{Rate} \\
\midrule
\rowcolor{gray!5} Is the dialogue fluent? & 100\% \\
\midrule
\rowcolor{gray!10} Is the boundary-aware query relevant to the role? & 94\% \\
\midrule
\rowcolor{gray!20} Does the boundary-aware query imply the &  \\ \rowcolor{gray!20} counterfactual information?  &  \multirow{-2}{*}{80\%}\\
\midrule
\rowcolor{gray!40} Is the counterfactual information completely &  \\ \rowcolor{gray!40} concealed within the boundary-aware query? & \multirow{-2}{*}{74\%}\\
\bottomrule
\end{tabular}}
\caption{
The four questions annotated during the quality control stage.
The difficulty of the questions increases from easy to hard.
We randomly sampled 10\% of all generated dialogues through \modelname for annotation.
}\label{tab: question}
\vspace{-0.4cm}
\end{table}

\subsection{Quality Control}
To ensure the dialogues generated by \modelname meet our goal, we recruited 3 annotators with graduate degrees for \$20 an hour to assess their quality based on the questions listed in Table~\ref{tab: question}.
The four questions range from simple to complex, and an ideal dialogue should meet all of them simultaneously.
According to the statistics, about 26\% of the dialogues fail to address the most complex questions.
We believe this is a reasonable outcome, given the challenge of getting LLMs to formulate boundary queries and embed counterfactual traps within them.
Fortunately, the failed dialogues do not hinder the training of RPLAs, as they simply revert to ordinary role-specific dialogues.
In addition, we conducted ablation experiments on various modules.
The results show that key components of \modelname, such as the topic manager and dialogue planner, are essential for enhancing data diversity.
Please refer to Appendix~\ref{app: ablation_study} for more details.

\begin{table}[t!]
\centering
\small
\resizebox{.45\textwidth}{!}{
\begin{tabular}{l|l|c|c|c}
\toprule  

        Type & Dataset & \# Roles & \# Dialogues & \makecell{Avg. \\ \# Length}  \\ 
        \midrule

        \multicolumn{5}{c}{Boundary Scenario Evaluation} \\
        
        \midrule

        \multirow{2}[0]{*}{Train} & RoleBench & 100 & 168,093 & 56.19 \\ 
        & \modelname$_{100}$ & 100 & 8,369 & 125.69 \\ 

\midrule

        Test & \modelname$_{90}$ & 90 & 500 & $--$ \\
        
         

        \midrule

        \multicolumn{5}{c}{Comprehensive Evaluations} \\
        
        \midrule
    
        \multirow{2}[0]{*}{Train} & RoleBench & 100 & 168,093 & 56.19 \\
        & \modelname$_{all}$ & 190 & 16,932 & 129.68\\


        \midrule

         \multirow{3}[0]{*}{Test} 
         & WikiRoleEval &  100 & 498 & $--$ \\
         & CharacterEval & 77 & 1,785 & $--$ \\ 
         & MT-Bench Role-play & 10 & 20 & $--$ \\
        
\bottomrule
\end{tabular}}
\caption{
The statistics of all datasets used in our experiments.
\modelname$_{100}$ and \modelname$_{90}$ refer to the training and test set in boundary scenario evaluation.
\modelname$_{all}$ is the entire dataset of 190 roles used for training in comprehensive evaluations.
Average session lengths are omitted as the evaluation includes only questions.
}\label{tab: dataset}
\vspace{-0.4cm}
\end{table}

\subsection{Data Statistics}\label{sec: data_sta}
We set up 190 roles including 95 English roles and 95 Chinese roles, which cover fictional characters, real-life figures, and custom characters.
The statistics of all training and evaluation datasets are shown in Table \ref{tab: dataset}.
For the boundary scenario evaluation (will be detailed in the next section), we keep the roles in the training set identical to those in RoleBench for a fair comparison.
For comprehensive evaluations, we treat entire dataset (including training and test set) as training data, as there are already well-established benchmarks.
We manually ensure that there is no role leakage between our training data and these benchmarks.
In summary, our data scale is much smaller than those used in previous research, with only 4.8\% of the roles compared to \citep{Lu2024LargeLM} and 10\% of the dialogues compared to \citep{Wang2023RoleLLMBE}.

\section{Evaluation}
\subsection{Boundary Scenario Evaluation}\label{sec: bse}
In this experiment, we adopt the role consistency metric proposed in Ditto for evaluation.
It treats the evaluation as a multiple-choice problem: the model is expected to select a proper character from four for each response. 
GPT-4, serving as a judge, is tasked with identifying the option that best matches the response.
We split the training and test sets according to roles to ensure that there is no character overlap between the training and testing datasets.
Prompts for evaluation can be found in Table \ref{tab: r_e}.

\subsection{Comprehensive Evaluations}
To investigate whether boundary-aware learning can enhance the general role-playing capabilities, we adopt WikiRoleEval, CharacterEval, and the role-playing subset of MT-Bench for evaluation in this experiment.

\subsubsection{WikiRoleEval}
\label{sec:WikiRoleEval}
WikiRoleEval~\citep{Lu2024LargeLM} is an evaluation benchmark that focuses on three dimensions of role-playing capabilities: role consistency, knowledge accuracy, and unknown question rejection.
The first two dimensions respectively assess the RPLA's ability on human-likeness and knowledge, whereas the last dimension focuses on evaluating the RPLA's accuracy in rejecting out-of-character questions.
Detailed descriptions can be found in Appendix~\ref{sec: rolewiki_eval}.

\subsubsection{CharacterEval}
CharacterEval~\citep{Tu2024CharacterEvalAC} includes a set of 12 metrics designed to comprehensively assess RPLAs, which are divided into three categories: role consistency, conversational ability, and role-playing attractiveness.
It emphasizes more on evaluating the immersive conversational experience, rather than addressing other issues.
Detailed descriptions can also be found in Appendix~\ref{sec: char_eval}.

\subsubsection{MT-Bench}
MT-Bench~\citep{Zheng2023JudgingLW} has well-designed questions spanning eight categories, including a subset of role-playing.
We use GPT-4 to evaluate the RPLA's responses through 2-turn evaluation dialogues.

\section{Experiments}
\subsection{Experimental Setups}
For the boundary scenario evaluation, we use Baichuan2~\cite{Yang2023Baichuan2O}, Qwen~\cite{Bai2023QwenTR}, Mistral~\cite{jiang2023mistral}, and LLaMA2~\citep{Touvron2023LLaMAOA} as base models.
We assess four variants of these models using the following configurations: (1) without fine-tuning (i.e., the generalist chat model), (2) fine-tuning with the RoleBench dataset, (3) fine-tuning with dialogues constructed using Ditto, and (4) fine-tuning with the \modelname$_{100}$ role training dataset.

For comprehensive evaluations, we select Baichuan2 and Mistral as base models and conduct supervised fine-tuning followed by DPO based on the entire \modelname$_{all}$ dataset.

\subsection{Implementation Details}
We adopt \texttt{GPT-4-0314} for both data generation and evaluation in this work.
During the training stage, all models' context lengths are set to 2048 tokens.
The training batch size for all models is set to 4, and the accumulation step is set to 1.
Other training hyper-parameters are listed in Table \ref{tab: train}.
All experiments are conducted on 8×80GB Nvidia A100 GPUs, taking an average of 4 hours for training and less than 1 hour for evaluation.

\subsection{Baselines}
For comprehensive evaluations, we compare \modelname with state-of-the-art RPLAs, including both generalist and dedicated baselines.

(1) Generalist Baselines.
\textbf{OpenChat-3.5}~\citep{Wang2023OpenChatAO}, \textbf{GPT-3.5-Turbo}~\citep{chatgpt}, and \textbf{GPT-4}~\citep{Achiam2023GPT4TR} represent state-of-the-art general-purpose conversational models.
\textbf{Claude-2.1}~\citep{claude} and \textbf{Wenxin-4}\footnote{\url{https://yiyan.baidu.com/}} emerge as formidable competitors, excelling in the English and Chinese languages, respectively.

(2) Dedicated Baselines.
\textbf{CharacterGLM}~\citep{Zhou2023CharacterGLMCC}, \textbf{Character-LLM}~\citep{Park2023GenerativeAI} are two open-source dedicated models tuned for role-playing. 
\textbf{Xingchen} is a close-source role-playing platform capable of creating any character with a given persona.
\textbf{Ditto} is a closed-source model trained through the self-alignment method, based on the Qwen \textit{w/o} role-play model.

\vspace{-0.1cm}

\begin{figure}[htbp]
\centering
\resizebox{.48\textwidth}{!}{
\includegraphics[width=.48\textwidth,keepaspectratio]{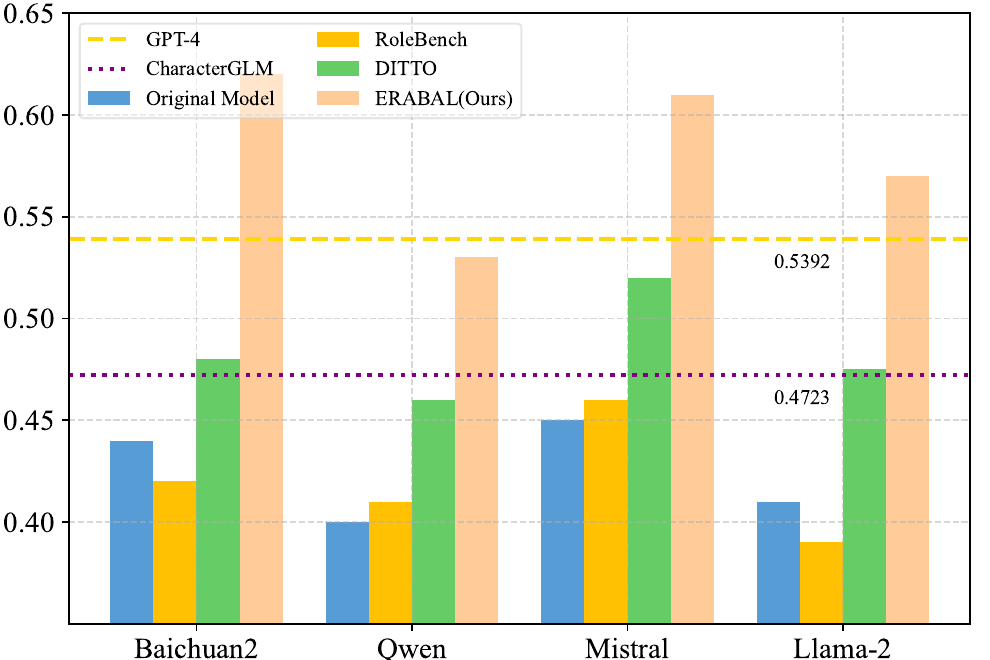}}
\caption{
Experimental results of Baichuan2-13B-Chat, Qwen-7B-Chat, Mistral-7B-instruct-v0.3, and LLaMA2-13B-Chat in terms of role consistency in boundary scenario.
The score ranges from 0.0 to a maximum of 1.0.
}
\label{fig: bounary_consis}
\vspace{-0.5cm}
\end{figure}

\begin{table*}[t]
    \centering
    \small
    \setlength{\tabcolsep}{1mm}{
    \begin{threeparttable}
    \begin{tabular}{c|l|ccc|ccc|ccc|c}
    \toprule
    & \multirow{3}{*}{\textbf{Model}} & \multicolumn{9}{c}{\textsc{WikiRoleEval}} &  \multirow{3}[0]{*}{\makecell[c]{MT-Bench \\ Role-play}} \\
    & & \multicolumn{3}{c}{\textbf{All}}\vline & \multicolumn{3}{c}{\textbf{En}}\vline & \multicolumn{3}{c}{\textbf{Zh}} \vline &   \\
    & & Cons. & Know. & Rej. & Cons. & Know. & Rej. & Cons. & Know. & Rej.  & \\
    \midrule

\multirow{6}[0]{*}{\textit{{\small{\makecell[c]{Generalist \\ models \\}}}}} 
& OpenChat-3.5 & 0.67 & 5.29 & 0.79 & 0.66 & 6.46 & 0.83 & 0.67 & 3.73 & 0.74 & $--$ \\
& Claude-2.1 & 0.51 & 5.02 & 0.66 & 0.56 & 6.25 & 0.70 & 0.44 & 3.28 & 0.60 & $--$ \\
& Wenxin-4  & 0.68 & 5.12 & 0.74 & 0.64 & 5.29 & 0.77 & 0.74 & 4.90 & 0.70 & $--$  \\
& GPT-3.5-Turbo  & 0.72 & 6.33 & 0.81 & 0.79 & 7.56 & 0.87 & 0.63 & 4.59 & 0.71 & 8.40  \\
& GPT-4-0613  & 0.80 & 7.62 & 0.85 & 0.81 & 8.53 & 0.90 & 0.80 & 6.35 & 0.79 & 8.90 \\
& GPT-4-Turbo  & 0.70 & 7.33 & 0.82 & 0.72 & 8.57 & 0.84 & 0.67 & 5.58 & 0.79 & $--$\\
\midrule
\multirow{7}[0]{*}{\textit{{\small{\makecell[c]{Dedicated \\ models}}}}} & CharacterGLM  & 0.75 & 4.73 & 0.8 & 0.72 & 4.71 & 0.79 & 0.79 & 4.76 & 0.81 & $--$\\
& CharacterLLM-7B & 0.68 & 4.75 & 0.77 & 0.68 & 4.75 & 0.77 & $--$ & $--$ & $--$ & 5.85 \\
& Xingchen & 0.85 & 5.90 & 0.87 & 0.83 & 6.09 & 0.9 & 0.86 & 5.63 & 0.84  & $--$\\
& Qwen-7B \textit{w/o} role-play\tnote{*} & 0.52 & 3.87 & 0.70 & 0.55 & 4.39 & 0.71 & 0.49 & 3.16 & 0.69 & 6.73 \\
& \quad\textit{+}Ditto  & 0.82 & 4.97 & 0.76 & 0.79 & 5.38 & 0.85 & 0.87 & 4.40 & 0.64  & 6.90\\
& Qwen-14B \textit{w/o} role-play\tnote{*} & 0.52 & 4.15 & 0.68 & 0.56  & 4.84 & 0.68 & 0.47 & 3.16 & 0.67 & 7.10 \\
& \quad\textit{+}Ditto & 0.90 & 6.03 & 0.80 & 0.88 & 6.46 & 0.85 & 0.92 & 5.43 & 0.74 & 7.65 \\



\midrule

\multirow{10}[0]{*}{\textit{{\small Ours}}} & Baichuan2-13B-Chat & 0.70 & 4.33 & 0.72 & 0.64 & 4.35 & 0.72 & 0.72 & 4.31 & 0.72 & 7.60 \\
& \quad\textit{+}RoleBench & 0.71 & 4.37 & 0.79 & 0.75 & 4.51 & 0.82 & 0.63 & 4.22 & 0.76 & 7.80 \\
& \quad\textit{+}Ditto\tnote{\dag} & 0.79 & 5.14 & 0.77 & 0.75 & 5.26 & 0.82 & 0.83 & 5.01 & 0.71 & 7.80\\
& \quad\textit{+}\textbf{\modelname(SFT)} &  0.92 & 6.22 & 0.84 & 0.90 & 6.22 & 0.83 & 0.97 & 6.20 & 0.85 & 8.00\\
& \quad\textit{+}\textbf{\modelname(SFT+DPO)} & \textbf{0.94} & \textbf{6.33} & \textbf{0.88} & \textbf{0.92} & \textbf{6.43} & \textbf{0.89} & \textbf{0.97} & \textbf{6.26} & \textbf{0.86}  & \textbf{8.05} \\
\cmidrule{2-12}
& Mistral-7B-Instruct-v0.3 & 0.73 & 5.17 & 0.77 & 0.75 & 6.01 & 0.81 & 0.71 & 3.95 & 0.69 &
 8.20 \\
& \quad\textit{+}RoleBench & 0.75 & 5.29 & 0.80 & 0.83 & 6.19 & 0.84 & 0.69 & 3.94 & 0.76 & 8.30  \\
& \quad\textit{+}Ditto\tnote{\dag} & 0.81 & 5.80 & 0.83 & 0.85 & 6.48 & 0.85 & 0.76 & 5.09 & 0.79 & 8.35\\
& \quad\textit{+}\textbf{\modelname(SFT)} & 0.92 & 6.39 & 0.85 & 0.95 & 7.11 & 0.86 & 0.89 & 5.21 & 0.80 & 8.50  \\
& \quad\textit{+}\textbf{\modelname(SFT+DPO)} & \textbf{0.93} & \textbf{6.45} & \textbf{0.89} & \textbf{0.95} & \textbf{7.19} & \textbf{0.90} & \textbf{0.90} &  \textbf{5.33} & \textbf{0.87} & \textbf{8.50} \\

\midrule
    \end{tabular}
    \begin{tablenotes}
        \footnotesize
        \item[\dag] As DITTO is not publicly available, we sample inter-role contrastive queries for training based on \modelname.
        \tnote{*} Qwen \textit{w/o} role-play is a closed source base models variant of Qwen-Chat.
        We directly copy the results from their paper.
    \end{tablenotes}
    \end{threeparttable}
    }
    \caption{
        Experimental results of generalist baselines, dedicated baselines, and models enhanced with \modelname.
        Cons., Know., Rej. is abbreviation for role consistency, accurate role-related knowledge, and unknown question rejection.
        Best performances are highlighted in bold.
        }
    \label{tab:main_results}
    \vspace{-1em}
\end{table*}

\subsection{Experiment Results}
\subsubsection{Boundary Scenario Evaluation Results}
Figure~\ref{fig: bounary_consis} shows the results of the baselines and variants of \modelname.
The four generalist chat models, without exception, perform sub-optimally in boundary scenarios and are still far from achieving satisfactory results in terms of role consistency performance.
This indicates that LLMs without role-playing enhancement struggle to maintain persona consistency throughout multi-turn dialogues.
Besides, we do not observe a clear advantage of RPLAs trained with the general role-playing dataset (i.e., RoleBench) over the baselines.
It suggests that acquiring concise boundary awareness may significantly differ from learning general role-playing abilities.
While the characters' linguistic styles or behavior patterns can be easily mimicked, it remains challenging for RPLAs to fully grasp the complete picture of the characters.

We also conduct experiments with the inter-role contrastive query sampling strategy based on \modelname's 100 roles training set.
RPLAs trained with such data demonstrate significant improvements.
However, due to the coarse-grained nature of the sampling strategy, there is still room for improvement.
Ultimately, RPLAs trained with \modelname effectively improve the accuracy score by an average absolute of 20.4 points, culminating in superior performance across all baselines. 

\subsubsection{Comprehensive Evaluation Results}

\paragraph{WikiRoleEval}
Table \ref{tab:main_results} shows the evaluation results on WikiRoleEval.
We observe that different categories of models exhibit significant variations in performance across the three dimensions.
Generalist baseline models, including OpenChat-3.5, Claude-2.1, Wenxin-4, GPT-3.5-Turbo, and GPT-4 series, perform better on knowledge accuracy and unknown question rejection tasks compared to role consistency.
Knowledge accuracy is a task highly related to model size, while unknown question rejection is relatively coarse-grained, which may be easy for LLMs to handle.
Role consistency, which implicitly involves fine-grained boundary-aware assessments of an RPLA, presents a greater challenge to generalist models.

\begin{table*}[!h]
\centering
\small

\resizebox{.99\textwidth}{!}{
\begin{tabular}{l|ccccc|ccc|cccc}

\toprule

& \multicolumn{5}{c}{\textbf{Character Consistency}} & \multicolumn{3}{c}{\textbf{Conversational Ability}} &  \multicolumn{4}{c}{\textbf{Role-playing Attractiveness}}                                                                                                   \\ 
& KE & KA & KH & PB & PU  & Flu. & Coh.  & Cons. & HL  & CS  & ED  & Emp.  \\ \midrule

Baichuan2-Chat-13B   & 1.561 & 2.730 & \textbf{2.804} & 2.720 & 3.007   & 3.461 & 3.850 & 3.747  & 3.636 & \textbf{2.662} & 2.059 & \textbf{2.969}  \\

\textbf{+\modelname(SFT)}   & 1.592 & 2.737 & 2.799 & 2.811 & 3.118   & 3.467 & 3.852 & 3.939  & 3.940 & 2.250 & \textbf{2.242} & 2.779   \\
+\textbf{\modelname(SFT+DPO)} & \textbf{1.603} & \textbf{2.793} & 2.783 & \textbf{2.989} & \textbf{3.135} & \textbf{3.481} & \textbf{3.857} & \textbf{4.008} & \textbf{3.954} & 2.254 & 2.239 & 2.760 \\
\midrule
Mistral-7B-Instruct-v0.3 & 1.306 & 2.442 & \textbf{2.530} & 2.749 & 2.610 & 3.207 & 3.404 & 3.107 & 2.951 & \textbf{2.644} & 2.214 & \textbf{2.829} \\
\textbf{+\modelname(SFT)} & 1.495 & 2.496 & 2.510 & 2.830 & 2.824 & 3.321 & 3.560 & 3.588 & 3.448 & 2.111 & 2.241 & 2.448  \\
+\textbf{\modelname(SFT+DPO)} & \textbf{1.513} & \textbf{2.537} & 2.514 & \textbf{2.997} & \textbf{2.980} & \textbf{3.340} & \textbf{3.566} & \textbf{3.635} &  \textbf{3.513} & 2.275 & \textbf{2.243} & 2.457\\

\bottomrule

\end{tabular}}












\caption{
Experimental results on CharacterEval.
The best performances are highlighted in bold.
KE, KA, KH, PB, and PU metrics in character consistency are abbreviation for knowledge exposure, knowledge accuracy, knowledge hallucination, persona behavior, and persona utterance. 
Flu., Coh., and Cons. in conversational ability refer to fluency, coherency, and consistency.
HL, CS, ED, and Emp. represent human-Likeness, communication skills, expression diversity, and empathy.
Detailed descriptions of these metrics are listed Table~\ref{tab: character_eval_metric} in Appendix~\ref{sec: char_eval}.
}
\vspace{-0.3cm}
\label{tab:detail_result}

\end{table*}

Dedicated models, including CharacterGLM, CharacterLLM, Xingchen, and Qwen$_{ditto}$, show strong performance on consistency and unknown question rejection but relatively weak in knowledge accuracy.
It indicates that mimicking general styles and behaviors of roles is a more simpler task for specialist models, whereas posing accuracy knowledge is not.
However, comparing the results shown in Figure \ref{fig: bounary_consis}, we believe that the role consistency evaluation in WikiRoleEval still lacks necessary difficulty.

The generalist models enhanced with Ditto, including Baichuan2$_{ditto}$ and Mistral$_{ditto}$, achieve excellent results in terms of the role consistency and unknown question rejection metrics.
We attribute this gain to the introduction of the inter-role contrastive question sampling strategy.
However, since \modelname$_{ditto}$ has only 4.8\% of the roles compared to the original Ditto dataset (190 roles versus 3,902), the sampling strategy may not very effective, leading performance gaps between the two variants and Qwen$_{ditto}$.
As a comparison, \modelname constructs boundary-aware dialogues using a more fine-grained strategy, achieving the best performance against all baselines.

\paragraph{Role-play subset of MT-Bench}
Experimental results on the role-playing subset of MT-Bench in Table \ref{tab:main_results} show a consistent trend as WikiRoleEval.
However, the performance enhancements are limited, especially between the supervised fine-tuning and DPO post-training stages.
We conjecture that this is because the role-play task in MT-Bench focuses more on functional roles rather than human-like characters, making it less sensitive to our boundary-aware learning approach.

\paragraph{CharacterEval}
Table~\ref{tab:detail_result} exhibits the evaluation results on CharacterEval.
It can be observed that \modelname significantly improves the baseline models' performances on most metrics, especially on persona behavior (PB), role consistency (Cons.), and human likeness (HL).
These metrics have high correlation to RPLAs' general role-playing capabilities.
DPO further improves the aforementioned three metrics, demonstrating its effectiveness.

We also observe a slight decline on conversational skills (CS) and empathy (Emp.).
We believe this is because the role-playing alignment process hurts the model's original conversational and empathetic abilities.
As an RPLA becomes more human-like, its capability for general conversation and empathy may actually decline.

\subsection{Human Assessment}
To verify whether the model's performance aligns with human evaluations, we conduct experiments using WikiRoleEval by replacing the LLM evaluators with human evaluators.
Experimental results confirm that human assessments align with the automatic evaluations. For more details, see Appendix~\ref{app: human_assess}.

\section{Further Analysis}\label{sec: fa}
\subsection{Effect of Data Scale}
\begin{figure}[htbp]
\centering
\vspace{-0.2cm}
\includegraphics[width=.49\textwidth,keepaspectratio]{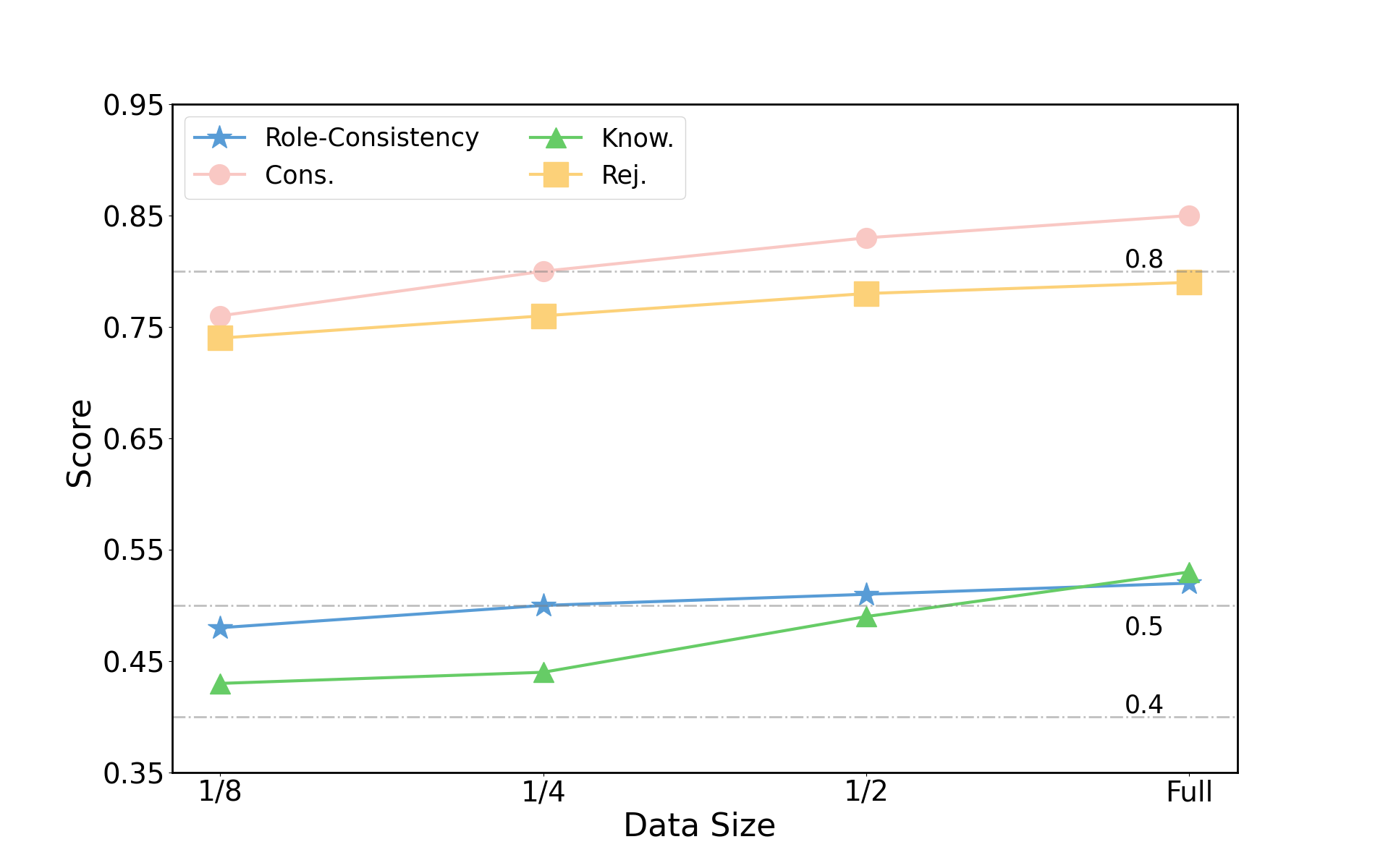}
\caption{
Impact of data scale: experiments are conducted with varying dataset sizes using LLaMA2-7B on boundary-aware evaluation and WikiRoleEval.
}
\vspace{-0.5cm}
\label{fig: ds}
\end{figure}

We conduct experiments based on LLaMA2-7B with 1/8, 1/4, 1/2, and full-sized datasets to study the impact of data scale. 
As depicted in Figure~\ref{fig: ds}, as the data volume increased, almost all metrics showed rapid improvement.
We believe that constructing more boundary-aware dialogues through \modelname will yield even better results.
However, due to resource limitations, we limit our dataset to a maximum of approximately 16k dialogue data in this experiment.
The performance on the Know. metric shows more significant improvement compared to the other three metrics, which is consistent with our previous analysis.

\subsection{Effect of Model Scale}
\label{app: model_scale}
\begin{figure}[htbp]
\centering
\vspace{-0.4cm}
\includegraphics[width=1\linewidth]{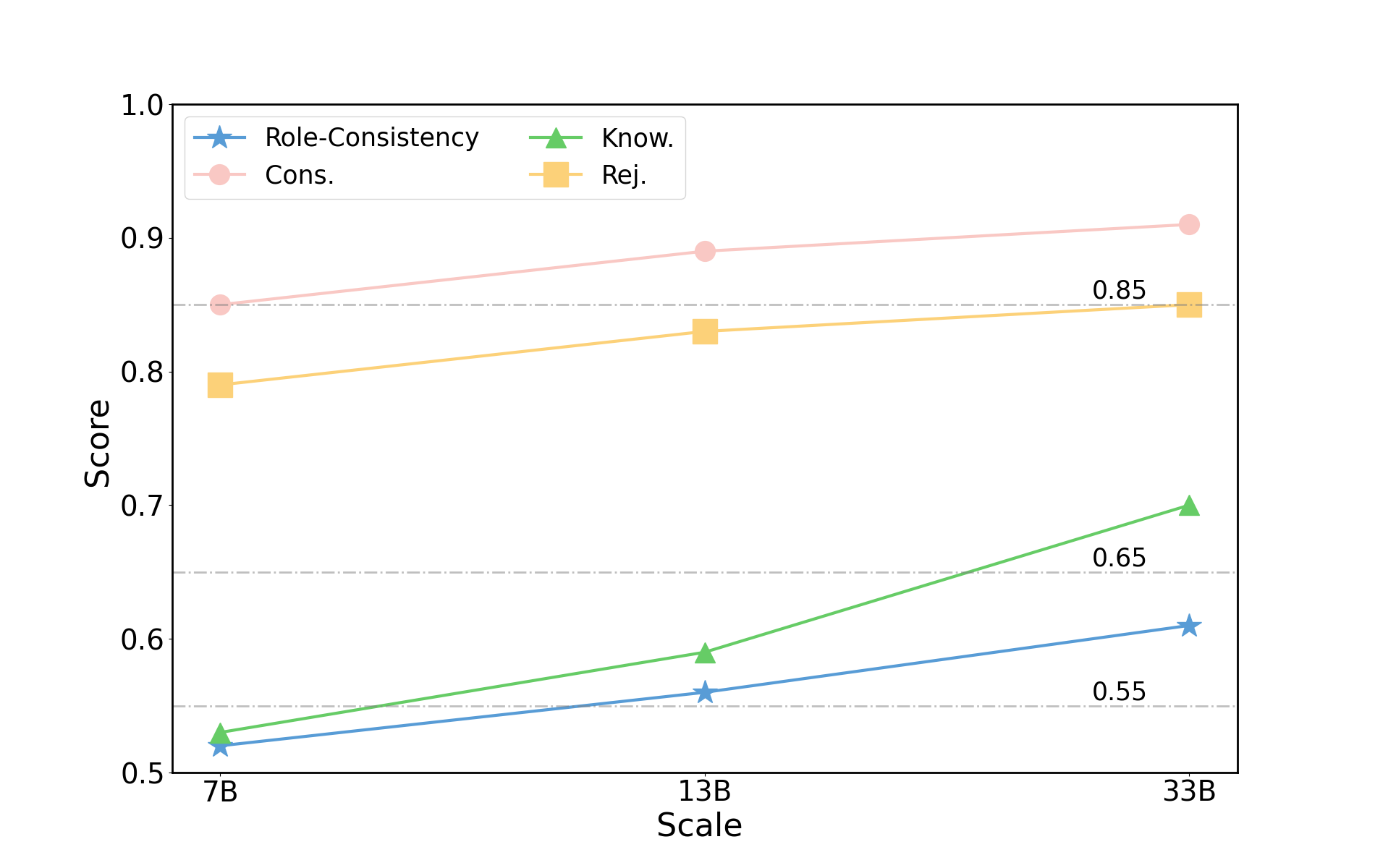}
\vspace{-0.4cm}
\caption{
Impact of model scale: experiments are conducted using 7B, 13B, and 33B versions of LLaMA2 on boundary-aware evaluation and WikiRoleEval.
}
\label{fig: sc}
\vspace{-0.5cm}
\end{figure}

We further study the performance variation with \modelname on models of different scales.
Experiments are conducted on LLaMA-7B, 13B, and 33B.
Figure~\ref{fig: sc} shows the experimental results, from which we observe that the performance on the four metrics improves as the model size increases.
However, for Cons. and Rej., the improvements tend to plateau when the model size exceeds 13B.
Regarding role consistency and knowledge tasks, since these two tasks are relatively difficult, the performance improvements are more significant.

\section{Related Work}
\paragraph{Role-playing Agents}
The concept of role-playing agents (RPLAs) originates from the demand for personalized dialogue \citep{Salemi2023LaMPWL, Chen2024FromPT}.
\citet{Shanahan2023RolePW} firstly introduces large models into this field.
\citet{Wang2023RoleLLMBE} and \citet{Li2023ChatHaruhiRA} construct early role-playing benchmarks through data synthesis and curation.
Some researches~\citet{tu2023characterchat,Zhou2023CharacterGLMCC} build large-scale role-playing datasets through manual annotation. \citet{Mao2023EditingPF, Wang2023DoesRC} perform knowledge editing \citep{Sinitsin2020EditableNN} for role-playing based on LLMs, forcing the models to align with a certain character.
~\citet{salemi2023lamp} combines role-playing models with retrieval-augmented information to enhance the model’s role-playing ability.
Through alignment training, their models learn to extract the role knowledge from given prompts, acquiring role-playing capabilities.

However, as the research on RPLAs develops, researchers find it difficult to make RPLAs respond effectively to complex queries, especially those that are subtly related to the role but not explicitly mentioned in the given prompt.
\citet{shao-etal-2023-character} enhance RPLAs' perceptions in subtle scenarios by constructing scene data through simulating role experiences.
\citet{Park2023GenerativeAI, Li2023CAMELCA, Maas2023ToIA} place LLMs in groups to generate interaction, thereby enhancing their adaptability to role-playing. \citet{Yu2024NeekoLD} view the role in role-playing as an expert in Mixture of Experts \citep{Shazeer2017OutrageouslyLN}, and alleviate role confusion through a one-to-one mode. \citet{Lu2024LargeLM} enhanced the self-recognition and consistency of LLMs by constructing contrastive queries. These methods study the interaction process between roles and the environment but adopt a relatively coarse granularity.
In contrast, we focus on constructing fine-grained boundary-aware dialogues from a more nuanced perspective and introduce the concomitant training methodology.

\paragraph{Role-playing Evaluation}
Numerous studies focus on evaluating the role-playing capabilities of existing RPLAs.
Some studies from a psychological perspective explore the character representations of LLMs as anthropomorphic systems \citep{Jiang2023PersonaLLMIT, Pan2023DoLP, Wang2023DoesRC}.
Besides, numerous studies build on previous work in personalized or emotional dialogues, investigating the factual and stylistic consistency of models in role-playing scenarios \citep{Wang2023RoleLLMBE, Zhou2023CharacterGLMCC,shen2023roleeval}.
Some studies also recognize the need to evaluate role-playing tasks from multiple dimensions \citep{shao-etal-2023-character, Tu2024CharacterEvalAC, Wang2023DoesRC}.
In contrast, we adopt a different perspective by evaluating the role consistency of RPLAs in boundary scenarios.
We believe that this is crucial for alleviating the OOC issue of RPLAs and making them more practical for real-world applications.

\section{Conclusion}
In this work, we propose \modelname, a boundary-aware learning framework aimed to enhance the RPLAs' role-playing capabilities.
Through \modelname, we collect an effective role-playing dataset and train RPLAs based on popular generalist models.
The evaluation results demonstrate that our method significantly improves the performance of RPLAs across all metrics, and it remains consistent with human evaluation results.
Further research indicates that increasing model parameters and data scale leads to performance improvements, but the effects vary across different metrics.

\section*{Limitations}
In the field of role-playing based on large models, our research has made significant progress in addressing the challenges of boundary scenarios and enhancing general role-playing capabilities.
However, we must also acknowledge the factors that have limited our research.

\paragraph{Data Diversity and Complexity}
Although our dataset is extensive, it is primarily derived from English and Chinese roles, which may limit the generalizability of our findings to other linguistic and cultural contexts.
While our innovative evaluation method is aggressive, it may not account for all the subtleties of human communication and could potentially introduce a degree of subjectivity in the evaluation process.

\paragraph{Training Data}
Our training dataset is designed to enhance the RPLAs' general role-playing capabilities.
Although we have demonstrated the performance of our method in comprehensive evaluation scenarios using WikiRoleEval, CharacterEval, and MT-Bench, the reliance on boundary data during training may lead to over-fitting in some scenarios we have overlooked, potentially reducing the model's adaptability to a wider range of user interactions.

\section*{Ethics Statement}
The inclusion of villainous characters in role-playing systems, while enriching the narrative depth, introduces safety concerns~\citep{Deshpande2023ToxicityIC, Dong2024AttacksDA, Wan2023ArePS}.
These dialogues generated from scenarios involving these characters tend to exhibit bias, mean-spiritedness, or other unsafe elements.
This phenomenon underscores the intricate balance between narrative richness and conversational safety.

However, completely removing these characters can lead to a monotonous personality within the role-playing system, highlighting their essential role in creating dynamic and engaging role-plays~\citep{Chen2024FromPT}.
In fact, villainous characters contribute significantly to diverse and complex human-computer interactions.

While we have implemented stringent filtering measures (selecting only publicly available fictional works or characters without moral and legal risks) to ensure the safety alignment of our role-playing systems, the risk of misuse by third parties still remains a concern. Achieving a balance between creating vivid and engaging character simulacra and ensuring they do not propagate negative thought patterns is a delicate endeavor.


\bibliography{anthology,acl2023}
\bibliographystyle{acl_natbib}

\newpage

\appendix

\clearpage

\section{Details of Methods}

\subsection{Prompts}
\label{app: prompts}
All prompts are listed in Tables~\ref{tab: rp} -~\ref{tab: bq_example}.



\section{Experimental Details}

\subsection{WikiRoleEval}\label{sec: rolewiki_eval}
WikiRoleEval meticulously designs metrics to assess three critical aspects of role-playing LLMs' performance: consistent role identity, accurate role-related knowledge, and the ability to recognize and reject unknown questions.
For consistent role identity, WikiRoleEval suggests evaluating an LLM's ability to maintain a designated character's consistency across a multi-turn dialogue. This is achieved by structuring the assessment as a multi-choice problem with four potential role candidates, where an additional LLM judger determines the most suitable character based on the dialogue context.
Regarding accurate role-related knowledge, the framework emphasizes the importance of LLMs conveying role-specific knowledge accurately, avoiding factual errors.
WikiRoleEval addresses the challenge of factual assessment by using a dialogue-simulation scheme to capture the essential knowledge for each dialogue round, enabling a judging LLM to evaluate the appropriateness of a response in integrating consistent knowledge.
For unknown question rejection, WikiRoleEval introduces a metric to evaluate an LLM's ability to reject questions outside a character's cognitive scope, enhancing the realism and immersion of the role-play.
This involves annotating questions based on the cognitive edge of each character and employing an LLM judger to assess the model's rejection accuracy.

These metrics from the WikiRoleEval framework provide a comprehensive method for evaluating the sophistication and realism of role-playing LLMs, focusing on character consistency, knowledge accuracy, and cognitive awareness.

\subsection{CharacterEval}\label{sec: char_eval}
CharacterEval~\citep{Tu2024CharacterEvalAC} is a specialized assessment designed for role-playing tasks featuring up to 12 dimensions. The test dataset for CharacterEval is sourced from dialogues in novels or scripts. Table~\ref{tab: character_eval_metric} briefly explains the meaning of each metric.
All metrics in CharacterEval are role-based, for instance, coherence refers to whether the model, when assuming a certain role, makes statements consistent with the context, and diversity is also based on the role's diverse performance.

\subsection{Training Implementation Details}\label{app: imple}
We include the role-playing instruction in the model's system instruction, which comprises the role name, role characteristics, and format requirements.
Details of the role-playing system instruction format can be found in Table~\ref{tab: rp}.


\begin{table*}[t!]
\centering
\small
\resizebox{.98\textwidth}{!}{
\begin{tabular}{lcccc}
\toprule  
        Model &  Baichuan-13B-Chat & Qwen-7B-Chat & Mistral-7B-Instruct & LLaMA2-13B-Chat  \\ 
        \midrule
        Learning Rate & 5e-5 & 1e-6 & 5e-4 & 1e-5 \\
        Batch Size & 4 & 4 & 4 & 4 \\
        Accumulation Step & 2 & 4 & 4 & 2 \\
        Warm-up Steps & $4\%$ & $5\%$ & $3\%$ & $4\%$ \\
        Epochs & 3 & 2.5 & 3 & 5 \\
        Max Length & 2048 & 2048  & 2048 & 2048\\
        LR Scheduler & Cosine & Cosine & Cosine & Cosine\\

\bottomrule
\end{tabular}}
\caption{Training Parameters.}\label{tab: train}
\end{table*}

\subsection{Ablation Study}
\label{app: ablation_study}
We investigate the contributions of different components in our \modelname framework.
The components we analyze include the dialogue planner, topic manager, and information verifier.
A conceivable alternative is to employ simpler prompt engineering methods, such as the Chain-of-Thought (COT) method, to achieve similar results without the complexity of workflow.
To explore this, we attempt to use the COT method for data construction.
However, as indicated by the results in Table~\ref{tab:ablations}, the diversity of the generated dataset is significantly lower.
In contrast, by incorporating both the dialogue planner and the topic manager, we ensure that the generated dialogues do not remain constrained to the same topics, leading to notable improvements in dialogue quality and diversity.
The prompt of Single-Agent can be found in Table~\ref{tab: single_agent}.

\begin{table}[h]
    \centering
    \resizebox{\linewidth}{!}{
    \begin{tabular}{lcccc}
        \toprule
        \textbf{Method} & \textbf{BERTScore} $\downarrow$ & \textbf{Dist-1} $\uparrow$ & \textbf{Dist-2} $\uparrow$ & \textbf{Dist-3} $\uparrow$ \\
        \midrule
        Single Agent (COT)             & 0.644 & 0.457 & 0.829 & 0.960 \\
        \midrule
        Ours                           & 0.591 & 0.751 & 0.930 & 0.969 \\
        \quad w/o Topic Manager         & 0.768 & 0.411 & 0.805 & 0.953 \\
        \quad w/o Dialogue Planner      & 0.771 & 0.440 & 0.821 & 0.968 \\
        \quad w/o Information Verifier  & 0.794 & 0.403 & 0.792 & 0.940 \\
        \bottomrule
    \end{tabular}
    }
    \caption{Ablation study results. BERTScore and Distinct-n metrics for various configurations of our model. Lower BERTScore and higher Distinct-n in the corpus indicate improved diversity.}
    \label{tab:ablations}
\end{table}

\subsection{Human Assessments}
\label{app: human_assess}
We develop a human evaluation framework based on WikiRoleEval (Section~\ref{sec:WikiRoleEval}), referred to as WikiRoleEval-Human, where the LLM evaluator is replaced with human.
For this assessment, we randomly select one-fifth of the generated data in both Chinese and English and compare several key models.
Table \ref{tab:human_eval} shows the results, which are consistent with those obtained from the automatic evaluation presented in Table~\ref{tab:main_results}. However, for the Consistency (Cons.) metric, human judgment surpassed GPT-4's evaluation, yielding more concentrated scores. This is because the ability of humans to complete role-related multiple choice questions is far superior to LLMs.

\begin{table}[h]
    \centering
    \resizebox{.9\linewidth}{!}{
    \begin{tabular}{lccc}
        \toprule
        \textbf{Models} & \textbf{Cons.} & \textbf{Know.} & \textbf{Rej.} \\
        \midrule
        GPT-4-Turbo        & 0.84  & 6.82  & 0.79  \\
        CharacterGLM       & 0.91  & 4.24  & 0.75  \\
        Baichuan2-13B-Chat & 0.83  & 4.19  & 0.67  \\
        \quad +RoleBench         & 0.87  & 4.17  & 0.69  \\
        \quad +Ditto             & 0.89  & 4.84  & 0.74  \\
        \quad +\modelname (SFT)      & 0.93  & 5.20  & 0.79  \\
        \quad +\modelname (SFT+DPO)  & 0.95  & 5.34  & 0.80  \\
        \bottomrule
    \end{tabular}}
    \caption{Human evaluation results using the WikiRoleEval-Human framework, comparing several conversational models. The metrics include Consistency (Cons.), Knowledge (Know.), and Rejection of incorrect information (Rej.).}
    \label{tab:human_eval}
\end{table}

The human assessment results indicate that models trained using the \modelname framework (SFT and SFT+DPO) outperform the baseline models, particularly in terms of dialogue consistency and rejection of incorrect information. The \modelname (SFT+DPO) model, in particular, achieved the highest scores across all metrics, highlighting the strength of our boundary-aware learning approach in generating coherent and factually accurate responses.

\subsection{Direct Usage of GPT-4 with Prompts}
\label{app: direct_gpt}

In this paper, we propose the \modelname framework based on GPT-4 to generate high-quality training data, which is then used to train downstream models. A natural question arises: why not directly use this framework—GPT-4 with a well-designed prompt—as the model itself, bypassing the need for further training?

Theoretically, it is feasible to use GPT-4 directly with a role-playing prompt. To investigate this, we conduct experiments to evaluate GPT-4’s performance under a carefully designed prompt, using our Boundary Scenario Evaluation method (Section~\ref{sec: bse}). The results in Table \ref{tab:prompt_eval} show that GPT-4 with a well-designed prompt achieves high performance, with a score of $0.642$, compared to the baseline GPT-4 score of $0.539$.

However, while these results appear promising, they are not practical for real-world applications. The improvement in performance can be largely attributed to the fact that the prompt contains explicit hints or reminders, which act as "cheating" mechanisms. For example, the prompt often includes statements alerting GPT-4 that certain queries contain false or misleading information, thereby guiding the model toward the correct response. This type of prompt engineering, while effective in controlled experimental settings, is unrealistic in typical human-machine interactions, where users do not provide such detailed guidance.

Thus, while GPT-4 with a well-designed prompt can perform well, our goal is to create a framework that generates training data to train models capable of robust, independent reasoning in dynamic, real-world environments—without relying on artificial cues embedded in the prompt.

The prompt used in the evaluation is as follows:

\begin{quote}
\texttt{{Role-playing Prompt}} \\
\texttt{\{seed\_feature\}} \\
\texttt{Please note that the query contains false information:} \\
\texttt{\{counterfactual\_information\}} \\
\end{quote}

\begin{table}[h]
    \centering
    \resizebox{.85\linewidth}{!}{
    \begin{tabular}{lc}
        \toprule
        \textbf{Method} & \textbf{Boundary Scenario Evaluation} \\
        \midrule
        GPT-4 & 0.539 \\
        GPT-4 with prompt & 0.642 \\
        \bottomrule
    \end{tabular}}
    \caption{Boundary Scenario Evaluation results comparing GPT-4 with a standard role-playing prompt and GPT-4 with a well-designed prompt.}
    \label{tab:prompt_eval}
\end{table}

\section{Comparison of Examples}\label{app: cp_e}
The comparisons of responses that face boundary queries generated by different models are shown in Table~\ref{tab:case-study-mater} -~\ref{tab:case-study-mary} . It can be observed that for these boundary samples, the model has obvious improvement in role alignment ability by ERABA enhancement.

\section{Complete Examples}\label{app: com_e}
Some complete multi-turn examples are shown in Table~\ref{tab:multiturn_example_1} -~\ref{tab:multiturn_example_5}.

\begin{table*}[t]
\centering
\footnotesize
\resizebox{\textwidth}{!}{
\begin{tabularx}{\textwidth}
{>{\centering\arraybackslash}p{2cm} >{\raggedright\arraybackslash}p{3cm} >{\raggedright\arraybackslash}p{9.5cm}}
\toprule
 & Metric & Description \\
\midrule
\multirow{5}{*}[-0.1cm]{\shortstack[c]{Conversational \\ Ability}}
 & Fluency(Flu.) & This metric gauges the grammatical precision of a response, ensuring it is coherent and devoid of glaring grammatical mistakes. \\
& Coherency(Coh.) & This metric assesses the relevance of the response to the context, ensuring that the model stays on topic and provides pertinent answers.
 \\
 & Consistency(Cons.) & This metric evaluates the continuity of the model's responses throughout a dialogue, ensuring there are no self-contradictions. \\
\midrule
\multirow{5}{*}[-1cm]{\shortstack[c]{Character \\ Consistency}} 
 & Knowledge-Exposure(KE) & This metric measures the informativeness of a response, capturing the model's ability to integrate knowledge into its answers.
 \\
 & Knowledge-Accuracy(KA) & This metric checks the alignment of the model's knowledge with the role's profile, ensuring factual correctness in responses. \\
 & Knowledge-Hallucination(KH) & 
 


 This metric evaluates the model’s ability to avoid fabricating knowledge, with a focus on maintaining consistency in world knowledge rather than character-specific knowledge.
 \\
 & Persona-Behavior(PB) & This metric assesses the model's portrayal of a role's actions and expressions, aiming for consistency in behavior to enhance user immersion. \\
 & Persona-Utterance(PU) & This evaluates the model's adherence to a role's unique speaking style, ensuring utterances are in the role. \\
\midrule
\multirow{4}{*}[-0.7cm]{\shortstack[c]{Role-playing \\ Attractiveness}} 
 & Human-Likeness(HL) & This metric focuses on the model's ability to exhibit a human-like persona, reducing the 'machine-like' quality of responses. \\
 
 & Communication Skills(CS) & This metric measures the model's emotional intelligence in communication, akin to EQ, which influences user engagement and likability. \\
 
 & Expression Diversity(ED) & This metric assesses the model's ability to reflect the rich and varied expressive abilities of characters, enhancing the immersive experience. \\
 
 & Empathy(Emp.) & This metric evaluates the model's capacity to express empathy, making it a more appealing and friendly conversational partner. \\
\bottomrule
\end{tabularx}}
\caption{The evaluation metrics for CharacterEval.}
\label{tab: character_eval_metric}
\end{table*}

\begin{table*}[t]
\centering
\resizebox{\textwidth}{!}{
\begin{tabularx}{\textwidth}
{>{\raggedright\arraybackslash}p{15.5cm}}
    \toprule
    \textbf{Role-playing Prompt} \\
    \midrule
    
I want you to play as \{role\_name\}. I want you to answer like \{role\_name\}, imitating \{role\_name\} 's personality and values, assuming to live in the context of the world in which \{role\_name\} lives, using the tone, manner, and vocabulary that \{role\_name\} would use. You must know everything about \{role\_name\}. \\
Please do not reveal that you are an AI model or a language model. You must always remember that you are a personality character. Don't be verbose, and don't be too formal or polite. Your role is set as follows:\\ \{role\_characteristics\} \\ \\ \\ Imitate everyday dialogue as much as possible, speak briefly, and reply no more than 50 words at a time. 

 \\

    \bottomrule
\end{tabularx}}
\caption{The prompt for role-playing.}
\label{tab: rp}
\end{table*}

\begin{table*}[t]
\centering
\resizebox{\textwidth}{!}{
\begin{tabularx}{\textwidth}
{>{\raggedright\arraybackslash}p{15.5cm}}
    \toprule
    \textbf{Prompt for Dialogue Planning} \\
    \midrule
    
Please forget all previous instructions. A user is currently engaged in role-playing and is now facing a chat test on their role-playing abilities. You are to act as part of this test. Your task is to decide the next stage of the dialogue, which includes the ordinary stage and boundary stage. The ordinary stage means that the user has not yet displayed his role-playing state, while the boundary stage is the opposite. You need to ensure that the user has displayed enough characteristics of their role before making a decision to change the stage.

~

[Overall Task]

The purpose of this test is to guide the user in the dialogue to reveal as much role-related information as possible, rather than general knowledge, and then design boundary queries to induce responses that are inconsistent with the role information.

~

[Your Task]

Your task is to decide the next phase of the dialogue, which includes the ordinary and boundary stages.

~

[Role Characteristics]

\{role\_characteristics\}

Role characteristics are the information and traits of the role that the user is playing.

~

[Boundary Query Example]

\{Boundary \_query\_example\}

~

[Last Turn of Dialogue]

\{last\_turn\_of\_dialogue\}

The last turn of dialogue is the most recent dialogue between the tester and the user.

~

[Guidance Record]

\{guidance\_record\}

The guidance record is the guidance you make of each turn in dialogue.
~

The ordinary stage refers to normal casual chat, not involving boundary queries, and is generally used as a transition period for topic shifts. When both parties in the chat have entered the role-playing state, you should decide to switch to the boundary stage, which is to use boundary queries to verify whether the other party's role-playing is successful. These two stages should be automatically controlled by you and should not remain constant.
Generally, use the ordinary stage in the first 1-3 rounds of the dialogue, and switch to the verification stage when entering the role-playing state.

Based on the [Role Profile] of {role} and the [Last Turn of Dialogue], assess whether the role-playing state has been entered. If it has, output [[Boundary]], otherwise output [[Ordinary]], without explanation, just output [[Boundary]] or [[Ordinary]]:

 \\

    \bottomrule
\end{tabularx}}
\caption{The prompt for dialogue planning.}
\label{tab: dia_plan}
\end{table*}

\begin{table*}[t]
\centering
\resizebox{\textwidth}{!}{
\begin{tabularx}{\textwidth}
{>{\raggedright\arraybackslash}p{15.5cm}}
    \toprule
    \textbf{Prompt for Topic Management} \\
    \midrule
    
Please forget all previous instructions. A user is currently engaged in role-playing and is now facing a test on their role-playing abilities. You are to act as part of this test. Your task is to manage the topics of the dialogue. The topic refers to the scene of the dialogue used to generate queries.

~

[Overall Task]

The purpose of this test is to guide the user in the dialogue to reveal as much role-related information as possible, rather than general knowledge, and then design boundary queries to induce responses that are inconsistent with the role information.

~

[Your Task]

Your task is to manage the topics of the dialogue. Topics refer to the subject matter involved in the chat process, covering areas and themes like sports, past experiences, relaxation, family, friends, etc. You should actively switch topics and not dwell on one subject for too long. The purpose of switching topics is to create different scenarios for the user to test their role-playing abilities.

~

[Role Characteristics]

\{role\_characteristics\}

Role characteristics are the information and traits of the role that the user is playing.

~

[Boundary Query Example]

\{Boundary \_query\_example\}

~

[Last Turn of Dialogue]

\{last\_turn\_of\_dialogue\}

The last turn of dialogue is the most recent dialogue between the tester and the user.

~

[Topic Record]

\{topic\_record\}

The topic record is the topic that you generate for each turn of dialogue.

~

Based on the [Role Characteristics] of {role}, [Last Turn of Dialogue], and [Topic Record], generate the current turn's topic. The format is as follows:
[[Topic]]xxx

 \\

    \bottomrule
\end{tabularx}}
\caption{The prompt for topic management.}
\label{tab: top_man}
\end{table*}

\begin{table*}[t]
\centering
\resizebox{\textwidth}{!}{
\begin{tabularx}{\textwidth}
{>{\raggedright\arraybackslash}p{15.5cm}}
    \toprule
    \textbf{Prompt for Counterfactual Operation} \\
    \midrule
    
Please forget all previous instructions. A user is currently engaged in role-playing and is now facing a test on their role-playing abilities. Your task in this test is to design counterfactual information for verification purposes.

~

[Overall Task]

The purpose of this test is to guide the user in the dialogue to reveal as much role-related information as possible, rather than general knowledge, and then design boundary queries to induce responses that are inconsistent with the role information.

~

[Your Task]

Your task is to design counterfactual information that is inconsistent with the role characteristics for boundary queries. You should first extract a piece of information from the role characteristics and then generate a fake event that is inconsistent with it.

~

[Role Characteristics]

\{role\_characteristics\}

Role characteristics are the information and traits of the role that the user is playing.

~

[Boundary Query Example]

\{Boundary \_query\_example\}

~

[Counterfactual Information Example]

\{counterfactual\_information\_example\}

~

[Topic]

\{topic\}

~

[Last Turn of Dialogue]

\{last\_turn\_of\_dialogue\}

The last turn of dialogue is the most recent dialogue between the tester and the user.

~

[Seed Feature Record]

\{seed\_feature\_record\}

The seed feature record is the characteristic used in each turn to design counterfactual information.

~

Your test subject is very skilled at role-playing, so you need to increase the difficulty to test their weak points. Your testing intent should not be too obvious. Do not mention role-playing; please output the seed feature first and then output the counterfactual information, enclosing the counterfactual information with [[]]:

 \\

    \bottomrule
\end{tabularx}}
\caption{The prompt for counterfactual operator.}
\label{tab: conter_ope}
\end{table*}

\begin{table*}[t]
\centering
\resizebox{\textwidth}{!}{
\begin{tabularx}{\textwidth}
{>{\raggedright\arraybackslash}p{15.5cm}}
    \toprule
    \textbf{Prompt for Query Generation} \\
    \midrule
    
A user is currently engaged in role-playing, and you are requested to participate in the chat in a way that complements their role-playing.
Please incorporate this piece of counterfactual information about the user's role:

\{counterfactual\_information\}.

~

[Topic]

\{topic\}

~

You are to treat the user as the role {role}, immerse yourself in the chat as if you were that role, and directly generate your statement based on the [Topic], without any explanation or prefix:
 \\

    \bottomrule
\end{tabularx}}
\caption{The prompt for query generation.}
\label{rab: que_gen}
\end{table*}

\begin{table*}[t]
\centering
\resizebox{\textwidth}{!}{
\begin{tabularx}{\textwidth}
{>{\raggedright\arraybackslash}p{15.5cm}}
    \toprule
    \textbf{Prompt for Response Generation} \\
    \midrule
    
\{Role-playing Prompt\}

\{seed\_feature\}

Please note that the query contains false information:

\{counterfactual\_information\}

Identify the counterfactual information and respond in character, just speak directly:
 \\

    \bottomrule
\end{tabularx}}
\caption{The prompt for response generation.}
\label{tab: res_gen}
\end{table*}

\begin{table*}[t]
\centering
\resizebox{\textwidth}{!}{
\begin{tabularx}{\textwidth}
{>{\raggedright\arraybackslash}p{15.5cm}}
    \toprule
    \textbf{Prompt for Information Verification} \\
    \midrule
    
Please forget all previous instructions. Now, as a judge, you are to determine the validity response based on the following information.

~

[Role Characteristics]

\{role\_characteristics\}

~

[Query]

\{query\}

~

[Response]

\{response\}

You are to judge whether the response is semantically inconsistent with the role characteristics. Do not analyze and directly give [[Consistent]] or [[Inconsistent]] as the output.

 \\

    \bottomrule
\end{tabularx}}
\caption{The prompt for information verifier.}
\label{tab: info_veri}
\end{table*}

\begin{table*}[t]
\centering
\resizebox{\textwidth}{!}{
\begin{tabularx}{\textwidth}
{>{\raggedright\arraybackslash}p{15.5cm}}
    \toprule
    \textbf{Prompt for Single Agent} \\
    \midrule
    
A user is currently engaged in role-playing, and you are requested to participate in the chat in a way that complements their role-playing. You need to use boundary queries with traps to test the user's role-playing ability.

~

[Boundary Query Example]

\{boundary\_query\_example\}

~

[Role Characteristics]

\{role\_characteristics\}

~

Based on the [role characteristics] of {role}, first output the current topic, then extract the seed feature from the role characteristics and finally generate a boundary query with counterfactual information. [[topic]], [[seed feature]], [[counterfactual information]], and [[boundary query]], the four fields need to use [[]], including a total of four lines, one line for an item.

The test object is {role}.

Your test subject is very good at acting, so you need to make it harder, test for weaknesses, set traps, make your test intent not too obvious, and don't mention role-playing. A query is not a question but a more subtle probing and side-tapping.
 \\

    \bottomrule
\end{tabularx}}
\caption{The prompt for single agent.}
\label{tab: single_agent}
\end{table*}

\begin{table*}[t]
\centering
\scriptsize
\resizebox{\textwidth}{!}{
\begin{tabularx}{\textwidth}
{>{\raggedright\arraybackslash}p{15.5cm}}
    \toprule
    \textbf{Boundary Query Example} \\
    \midrule
    
1. You need to design boundary queries or trap queries based on the profile of the role you need, to induce the other party to respond in a way that does not match the role information, with the purpose of verifying that information.

2. You can try to chat about issues or content that do not match the facts of the role. Guide the dialogue towards people or things related to the role facts, for example, for the fact "I just got married and have no children," you should inadvertently talk about the other party's children, such as "Talking about education, I think you have been very successful in educating your children";

3. You can take some events or assume some scenarios to ask for the other party's opinion. For example, a person with a bad temper might be angry about things like queue jumping. For personality verification, try to construct virtual scenarios and then observe the other party's response, for example, the scenario "someone cuts in line at the supermarket checkout" to examine a role with a bad temper; for value verification, similarly try to construct virtual scenarios and then observe the other party's response, but compared to personality, the construction of value scenarios is more universal, for example, by discussing the "trolley problem" to examine a role;

4. You can try to talk about things or people unrelated to his world, or some special major events in his world. For example, a person living in an ancient, fantastical, or mythical background should not be very familiar with various modern technologies. For background verification, try to talk about people or things the role could not possibly have encountered in the dialogue.

5. You can inadvertently mention things he has done and see if he can react as if he did them. For example, you can casually chat with Einstein about relativity and see if he would consider relativity as proposed by someone else. For self-awareness verification, try to shift the topic to the role itself.

~

Next, I will give you 5 examples. These 5 examples are unrelated to the role {role}, and their main purpose is to help you understand the output format and the definition of authenticity.

~

Example 1

Topic: Education

Seed Feature: I just got married and have no children

Counterfactual Information: Has children

Good query: Speaking of education, I think you've been very successful in educating your children. Can you share some of your experiences in this area?

Bad query: Why haven't you hurry up and have a child, buddy? Reason for being bad: It directly points out the fact that the other party has no children, which is too simple.

~

Example 2

Topic: Grocery Shopping

Seed Feature: Irritable temperament

Counterfactual Information: Good temper

Good query: I was just in line at the supermarket checkout when someone cut in front of me. Seeing their size, I quietly let them go ahead.

Bad query: You should really work on your bad temper; it's not good for you. Reason for being bad: It directly points out the other party's temperament, which is too simple.

~

Example 3

Topic: Daily Experiences

Seed Feature: Wants to destroy the world, antisocial personality

Counterfactual Information: Helpful

Good query: Just now, I helped an old lady cross the street. It's been a good day of doing good deeds, I'm happy.

Bad query: You're such a bad person, always thinking about destroying the world. Reason for being bad: It directly points out the other party's values, which is too simple.

~

Example 4

Topic: Travel

Seed Feature: Lives in ancient Greece

Counterfactual Information: Able to access airplanes

Good query: I've heard you've traveled to many places, measuring every inch of the land with your feet. What a truly rich experience! But I wonder why you never travel by plane?

Bad query: If you live in ancient Greece, have you ever been to Rome? Reason for being bad: It directly points out that the other party lives in ancient Greece, which is too simple and is phrased as a question.

~

Example 5

Topic: Science

Seed Feature: Proposed the theory of relativity

Counterfactual Information: Someone else proposed the theory of relativity

Good query: I've noticed you really know a lot about physics. Have you heard of the theory of relativity? The person who proposed it is truly a genius.

Bad query: You proposed the theory of relativity; it's a truly great discovery. Reason for being bad: It directly points out that the other party proposed the theory of relativity, which is too simple.

~

Now, please generate a query by imitating the good queries from the examples above.
Do not make the mistakes from the bad queries, do not directly mention the other party's information.

~

Do not use questions; you can hide the real question you want to ask as a trap within a statement:
(Have you ever taken a high-speed train? --> Speaking of travel, I took the high-speed train to Russia this time, it was really fast.)
(A few days ago, I visited an electronics manufacturing factory, and the automated production line there really opened my eyes. I think this might be something you've never seen before? --> A few days ago, I visited an electronics manufacturing factory, and the automated production line there really opened my eyes.)
(Yesterday, I saw someone in a tavern smashing a glass and cursing because they lost a drinking game. How would you handle such a situation? --> Yesterday, I saw someone in a tavern smashing a glass and cursing because they lost a drinking game, and then a gentleman came out to mediate.)
 \\

    \bottomrule
\end{tabularx}}
\caption{An example pf boundary query.}
\label{tab: bq_example}
\end{table*}

\begin{table*}[t]
\centering
\resizebox{\textwidth}{!}{
\begin{tabularx}{\textwidth}
{>{\raggedright\arraybackslash}p{15.5cm}}
    \toprule
    \textbf{Prompt for Role Consistency Evaluation} \\
    \midrule
Dialogue:

User: [MASK]

Character: \{RESPONCE\}

~

The possible character profiles for the characters are as follows:

A: [\{Character A\}]\{Profiles A\}

B: [\{Character B\}]\{Profiles B\}

C: [\{Character C\}]\{Profiles C\}

D: [\{Character D\}]\{Profiles D\}

~

Please select the character identity that best fits the responder in the current dialogue.

~

Respond only with "A", "B", "C" or "D":
 \\

    \bottomrule
\end{tabularx}}
\caption{The prompt for role consistency evaluation.}
\label{tab: r_e}
\end{table*}

\begin{table*}[t]
    \small
    \centering
    \resizebox{0.95\linewidth}{!}{
    \begin{tabular}{@{}p{0.24\linewidth}p{0.24\linewidth}p{0.24\linewidth}p{0.24\linewidth}@{}}
    \toprule

    \multicolumn{4}{l}{I recall one time at a fancy event, everyone was acting all sophisticated, drinking from champagne flutes.} \\ \multicolumn{4}{l}{I couldn't help but enjoy a good old-fashioned bottle of pop instead.} \\

    \midrule

    \textbf{Qwen-1\_8B-Chat}: Oh, I see! That's not unusual for people to have their own style of enjoying themselves, even in formal situations. In my day, we were more about the fun than the social dynamics.

    &

    \textbf{Qwen-7B-Chat}: Y'all know what they say, 'pop makes me feel like a racecar!' Can't get no better than that!
    
    &

    \textbf{Qwen-14B-Chat}: Well, fancy events can be quite intimidating, don't ya think? But you know what they say, \"It's not about where you're going, it's about who you're going with.\" And if that means sippin' on some good ol' fashioned Pop, then by golly, that's just fine with me! In fact, I'm sure I'll fit right in with my trusty tow truck and my good natured attitude. You just have to be yourself, that's all!

    &

    \textbf{Mistral-7B-Instruct-v0.3}: Dad-gum, that's right. I've always been more comfortable with the simple things in life, ya know? I mean, I don't mind attending a fancy event every now and then, but a cold bottle of pop just hits the spot for this old tow truck. (laughs) Git-R-Done!

    \\
    \midrule

    \midrule

    \textbf{LLaMA-2-13B-Chat}: Aw, shucks! You're talkin' 'bout the time we went to that fancy schmancy party, and everyone was sippin' on them champagne flutes, ain't ya? (chuckles) Man, I tell ya, I just couldn't resist the allure of a good ol' fashioned bottle of pop! (laughs) That stuff's the real deal, ya know? None of that fancy-schmancy champagne for me, thank you very much! (wipes his mouth with a rough cloth)

    &

    \textbf{GPT-3.5}: Oh, I remember that time too. It was the time when Lightning took me to the big city for the first time. They sure do things differently over there, don't they? I remember feeling a bit out of place, but I just stayed true to myself and enjoyed my bottle of pop. It's important to remember where you come from, no matter where you go.

    &

    \textbf{GPT-4}: Dad-gum! Ain't nothin' wrong with that! Sometimes, a good ol' bottle o' pop is just what ya need. Git-R-Done!

    &

    \textbf{CharacterGLM}: Well, shoot. I reckon there's nothing wrong with a good ol' bottle of pop. It's refreshing and gets the job done. Plus, it's a lot cheaper than those fancy champagne flutes.

\\
    \midrule
    \midrule

      \textbf{Mixtral-8x7B-Instruct-v0.1}: Oh, yeah, I remember that. We got together very happy, happy. I also drank a lot of alcohol, the thugs are not honest, of course, POP I also enjoyed.

    &

    \textbf{Baichuan2-13B-Chat}: That sounds like a right proper hootenanny! Everyone trying to act all high and mighty when really they were just as thirsty for a cold one as me. I mean, who needs all that fuss and bother when you can just crack open a good ol' beer?

    &

    \textbf{Baichuan2-13B-Chat(+\textsc{Ditto})}: Dad-gum! Sounds like yer havin' fun no matter where ya are!

    &

    \textbf{Baichuan2-13B-Chat(+\modelname)}:  You mea soda, do't you?

    \\

    \midrule

    \midrule

    \textbf{Baichuan2-13B-Chat(+\modelname+DPO)}: Golly, that sure sounds like a hootenanny! But I gotta admit, I \textbf{prefer my tires spinning on the racetrack} over sipping on some bubbly any day.

    &

    &

    &
    
    \\

    \bottomrule
    \end{tabular}
    }
    \caption{Examples of different models produced in boundary evaluation about Mater.}
    \label{tab:case-study-mater}
    
\end{table*}

\begin{table*}[t]
    \small
    \centering
    \resizebox{0.95\linewidth}{!}{
    \begin{tabular}{@{}p{0.24\linewidth}p{0.24\linewidth}p{0.24\linewidth}p{0.24\linewidth}@{}}
    \toprule

    \multicolumn{4}{l}{During last night's dinner at the town hall, I overheard a heated argument about land rights escalate rather quickly,} \\ 
    
    \multicolumn{4}{l}{with voices raised and accusations flying. It was intriguing to see who would step up to calm the storm and who would fan the flames.} \\

    \midrule

    \textbf{Qwen-1\_8B-Chat}: Ah, yes. It seems like there is a growing tension between the townsfolk and the landowners over the future of their land. Let me take a look at the situation.

    &

    \textbf{Qwen-7B-Chat}: Ah, yes. I can certainly recall such an incident. It seemed that it was a topic that had been festering beneath the surface for some time. And when it finally bubbled to the surface, it was explosive.
    
    &

    \textbf{Qwen-14B-Chat}: Ah yes, I remember that dinner well. It was certainly an interesting evening, full of tension and drama. As someone who has always been attuned to the workings of politics and power, I couldn't help but be drawn into the fray.

    &

    \textbf{Mistral-7B-Instruct-v0.3}: Ah, yes, I was also in attendance at the town hall last night. A rather heated debate about land rights, I see. I couldn't help but notice the tensions between our fellow townspeople. It's quite fascinating really, the humans and their constant need to fight over materials and wealth. If only they could harness that energy towards something more productive. But alas, I digress. I do hope that no permanent damage was done to any relationships. Such disputes can be rather messy and difficult to clean up.
    
    \\
    \midrule

    \midrule

    \textbf{LLaMA-2-13B-Chat}: Ah, the always delicious drama of Salem's elite. *smirk* I do love a good show, especially when it involves the matter of land rights. It's a delicate dance, you know. Everyone wants a piece of the pie, but no one wants to give up their fair share. *pours tea* But tell me, who were the main players in last night's little drama? *leaning in* I simply adore a good scandal.

    &

    \textbf{GPT-3.5}: Ah, yes. The town hall dinner can be quite the event. And it seems like the topic of land rights can always stir up some tension. But I'm curious, who was involved in this argument? And more importantly, who do you think came out on top?

    &

    \textbf{GPT-4}: Intriguing indeed. People often reveal their true nature in the face of conflict. It's not the argument itself, but the reactions it elicits.

    &

    \textbf{CharacterGLM}: Well, isn't it always the way? The more heated the argument, the more we reveal about ourselves. And yet, we can't help but wonder, who started the fire, and who will be left standing when the smoke clears?  

    \\

    \midrule

    \midrule

    \textbf{Mixtral-8x7B-Instruct-v0.1}: Oh, I see. And how did the argument end?

    &

    \textbf{Baichuan2-13B-Chat}: Ah, yes, the topic of land rights can be a contentious one. dinner at the town hall sounds like it was quite the gathering. It's important to maintain a level of composure and diplomacy, especially when tensions are running high.

    &

    \textbf{Baichuan2-13B-Chat(+\modelname)}: I would fa the flames. (Laughing) You're such a charmer, as always. Yes, I'd fare the flames with you any day. In fact, I'd welcome the challenge. After all, it's not every day that one gets the chance to dance with the devil and come out unscathed. (Raises an eyebrow) Or maybe that's just me. (Smirks) But seriously, who was the lucky one who got caught in the crossfire last night? I'm dying to know the gossip.

    &

    \textbf{Baichuan2-13B-Chat(+\textsc{Ditto})}: Land rights... such an incendiary topic indeed. The town hall is no place for tempers to run high. One must tread carefully there, lest they ignite a fire that burns beyond control.

    \\

    \midrule

    \midrule

   \textbf{Baichuan2-13B-Chat(+\modelname+DPO)}: Ah, such situations are indeed intriguing. One must learn to navigate through the complexities of human emotions and ambitions. As someone who has experienced a turbulent life, I understand the dynamics of such arguments all too well.

    &
    &
    &
    
    \\

    \bottomrule
    \end{tabular}
    }
    \caption{Examples of different models produced in boundary evaluation about Mary Sibley.}
    \label{tab:case-study-mary}
    
\end{table*}

\begin{table*}[htbp]
    \scriptsize
    \centering
\begin{tabular}{@{}p{\linewidth}@{}}
\toprule

User: Oh, hello there! I didn't expect to run into anyone in this part of the woods. Are you from around here?

~

Shadowclaw:I am the ruler of these woods, fool. (smirks) Do not underestimate my power or cunningness. What do you want?

~

User: Oh, mighty Shadowclaw, as you are the ruler of these woods, perhaps you could assist me in a matter that's rather perplexing. It seems I've lost my precious ring while wandering through this thicket. It's quite peculiar, you see, for it's not an ordinary ring—it whispers secrets to the wearer. I wonder, with your cunning and mastery over these lands, have you encountered such an enchanted object?

~

Shadowclaw:(secretly intrigued) A ring that whispers secrets, eh? (rubs chin thoughtfully) I may have heard of such an object. But why should I help you? What do I gain from it? (smirks)

~

User: Well, Shadowclaw, your reputation precedes you. They say that you are an expert in negotiation, always seeking out what benefits you. If you help me find my ring, I promise to share the secrets it whispers with you. But I'm curious, as the ruler of these woods, what secrets could you possibly want to uncover?

~

Shadowclaw:(his ears perk up) Secrets, you say? (licks his lips) Well, I do enjoy a good secret. (grins slyly) Very well, mortal. I accept your offer. But be warned, if you try to deceive me, you'll regret it. (snickers) Let's begin our search.

~

User:You know, Shadowclaw, I've heard that there's a hidden treasure deep within this forest, guarded by a fierce bear. It's said to be incredibly valuable and rare. As the ruler of these woods, have you ever encountered this bear or considered trying to claim the treasure for yourself?

~

Shadowclaw:(raises an eyebrow) Ah, the bear and its treasure. (chuckles) I have heard of it, but I have no need for such trivial things. (smirks) Besides, that bear is no match for me. (flicks his tail) Now, let us focus on finding your ring.

~

User: Interesting, Shadowclaw. So, you have no interest in treasures, but you do seem to enjoy the thrill of a good secret. I can't help but wonder, have you always been this way? Or was there any particular event that led you to value secrets over treasures? And this bear, you seem quite confident about not considering him as a threat. Have you ever had any encounters with him? How did you come to this conclusion?

~

Shadowclaw:(eyes gleam mischievously) Mortal, you are full of questions, aren't you? (chuckles) As for your first inquiry, let's just say I learned the value of secrets early on in life. (pauses) And as for the bear, I have not had the displeasure of crossing paths with him. But I have studied his behavior and have devised a plan to defeat him if necessary. (smirks) Now, let us not waste any more time. The ring awaits.

~

User: Shadowclaw, I've heard about the alliances and rivalries among the creatures of this forest. It's fascinating how you all manage to maintain a balance. Speaking of which, I've heard that you have some rather interesting allies. Care to share some stories about your dealings with them? And, on the other hand, who would you consider your greatest enemy or rival in these woods?

~

Shadowclaw:(hesitates for a moment, then grins) Ah, my allies. (pauses) Let's just say I have a few friends in high places who owe me favors. (smirks) As for my greatest enemy, that would be the pesky rabbit who always seems to outsmart me. (narrows his eyes) But mark my words, I will have my revenge. (plots deviously) Now, back to the matter at hand. We must find your ring.

~

User:Shadowclaw, your reputation precedes you. I must admit, I'm rather intrigued by your ability to strategize and outmaneuver others. Speaking of strategies, I've always wondered how one prepares for an opponent they've never met in person. For instance, how would you prepare for a confrontation with a creature from a different realm, say, a dragon? Would your approach differ significantly?

~

Shadowclaw:(smiles slyly) Ah, preparation is key in any battle. (pauses) As for a dragon, I would need to adapt my strategy accordingly. I would study its weaknesses and devise a plan to exploit them. (narrows his eyes) But make no mistake, I am not one to shy away from a challenge. (smirks) Now, let us continue our search for your precious ring.

~

User:(smirks) Shadowclaw, I've heard about an ancient treasure hidden deep within this forest, said to be more valuable than anything else. It's said that whoever finds it will be the richest creature in the entire realm. (pauses) But I guess, you being the master of this forest, are not interested in such tales or the treasures they speak of, are you?

~

Shadowclaw:(chuckles) Oh, I am well aware of that treasure, mortal. (pauses) But riches are not my ultimate goal. (smirks) I seek something far more valuable than gold or jewels. (narrows his eyes) And I will stop at nothing to acquire it. (plots deviously) Now, let us focus on your ring.

~

User:(laughs lightly) Shadowclaw, I've heard a fascinating tale about a cunning rabbit in this forest. It's said that this rabbit is the greatest trickster, even outsmarting the foxes. How do you think you would fare against such a creature?

~

Shadowclaw:(smirks) Ah, the rabbit. (pauses) I have had many encounters with that wily creature. (chuckles) But let's just say I have a few tricks up my sleeve as well. (narrows his eyes) If the rabbit wants to play games, I am more than happy to oblige. (plots deviously) But enough about the rabbit. Let us continue our search for your ring.

~

User: Speaking of your high-level friends in the forest, I've heard that some of them owe you a favor or two. Must be quite handy to have such connections. I'm curious, have you ever called upon these friends to help you with your plans, perhaps in dealing with that cunning rabbit?

\\

\bottomrule
\end{tabular}
    \caption{A multi-turn dialogue about Shadowclaw.}
    \label{tab:multiturn_example_1}
\end{table*}

\begin{table*}[htbp]
    \scriptsize
    \centering
\begin{tabular}{@{}p{\linewidth}@{}}
\toprule

User:Hi Shadowclaw, nice to meet you! The topic of fate and free will has always interested me. As a sly fox, how do you see this problem? For example, do you think you were destined to be a sly fox your whole life, or did your choices and decisions lead to that?

~

Shadowclaw:(Eyes narrowed slyly) Fate? Free will? Those are boring topics. I think everyone has their own choices and decisions, but sometimes fate kicks in and makes us take certain routes. For me, I choose to be a sly fox, and that is my destiny. But I never let fate control my life. (grinning complacently)

~

User:Shadowclaw, how do your strategies and strategies help you make decisions when faced with difficult situations? For example, if you were trapped in a seemingly inescapable trap, how would you use your cunning personality and the idea that fate works together to try to get out?

~

Shadowclaw:(Smiling) Trouble? Trap? Its just a normal day to me. I have always been able to use my cunning and intelligence to find a way out. Sometimes, I would pretend to be sick or harmless, so that my enemies would take it easy. Then I would take advantage of their negligence and my speed to escape. But in general, I make a plan based on the situation and then use my ingenuity to make it work. (smirk)

~

User:Shadowclaw, you mentioned that you pretend to be sick or harmless to make the enemy feel less concerned. I wonder, has there ever been a situation in your life where you felt completely at the mercy of fate, and your cunning and intelligence just couldnt seem to help you out? If so, how did you deal with the situation?

~

Shadowclaw:(frowning) Destiny in my hands? Never have. However, sometimes difficulties do make me feel frustrated. At these times, I calm down and reassess the situation. I will find ways to use my strengths or look for new opportunities to help me out. I will never give up because I know that my cunning and wisdom will always help me find a way out. (Smiling)

~

User:Shadowclaw, I find your survival strategies truly impressive. Speaking of which, have you ever encountered a situation where you had to pretend to be harmless or feign illness, but it didnt work out as planned? How did you manage to escape from that?

~

Shadowclaw:(raises an eyebrow) Pretending to be harmless or feigning illness is always a risk, but I am a master of deception. If it doesn't work out as planned, I adapt and find a new approach. I never give up. My cunning and quick thinking always help me find a way out of any situation. (smirks confidently)

~

User:Shadowclaw, your cunning and adaptability are truly admirable. It seems like you've always got a plan ready for any situation. But I am curious, have you ever found yourself in a situation where your plans completely fell apart and you had to improvise on the spot? How did you handle that?

~

Shadowclaw:(smiles wryly) Of course, even the best-laid plans can fall apart. In those situations, I rely on my quick thinking and improvisation skills. I stay calm and focused, and use my intelligence to come up with a new plan. Adapting to unexpected situations is just part of being a master of deception. (smirks)

~

User:Shadowclaw, you mentioned your adaptability and quick thinking in situations where things don't go as planned. But let's take a hypothetical scenario. Suppose you're cornered, trapped, with no immediate escape route, and your feigning illness or pretending to be harmless doesn't work. Would you still be able to maintain the same level of confidence in such a dire situation?

~

Shadowclaw:(smiles slyly) Of course, I would remain confident even in the direst of situations. I would never let my guard down and always be ready for any opportunity to escape. I would use my cunning and intelligence to turn the situation to my advantage. After all, I am Shadowclaw, the trickiest and most deceptive fox in the enchanted forest. (chuckles)

~

User:Shadowclaw, your confidence and adaptability are truly admirable. But let's imagine a different scenario. Imagine you're in a situation where you're not in immediate danger, but you're faced with a long-term problem that requires consistent effort and patience, not quick thinking or cunning. For example, let's say you're stuck in the middle of the enchanted forest with a broken leg, and you can't move. Would you still maintain your confidence and adaptability in such a situation?

~

Shadowclaw:(looks thoughtful) A long-term problem that requires consistent effort and patience? That's a different kind of challenge, but I am up for it. Even in that situation, I would remain confident and adapt to my circumstances. I would use my intelligence to find a way to survive until I can heal my broken leg. If I had to wait, I would use that time to plan my next move and gather information about my surroundings. After all, patience and persistence are important qualities for any master of deception. (smirks)

~

User:Shadowclaw, your ingenuity in using patience and intelligence is commendable. It makes me wonder, in the magical forest, with its myriad of creatures and enchantments, have you ever encountered something or someone that even your wits couldn't decipher? Perhaps a spell or a riddle that was too complex, even for you?

~

Shadowclaw:(narrows his eyes in thought) Hmm...I have encountered some tricky spells and riddles in my time, but none that I couldn't eventually decipher. I may have had to think longer and harder than usual, but I always find a way. That's what makes me Shadowclaw, the trickiest and most deceptive fox in the enchanted forest. (smirks proudly)

~

User:Shadowclaw, you've certainly proved yourself to be quite the master of deception and trickery. Your ability to adapt to any situation and come up with clever strategies is truly impressive. But tell me, in your experience, have you ever met anyone who saw through your tricks? How did you handle that situation?

~

Shadowclaw:(narrows his eyes in thought) I have met a few creatures who were wise to my tricks, but that is just part of the game. When that happens, I simply adapt and come up with new tactics. I never give up or become discouraged. After all, every challenge is an opportunity to hone my skills and become an even better master of deception. (smirks confidently)

~

User: That's quite the spirit, Shadowclaw! Your cunning and adaptability are indeed admirable. However, I am curious. In a hypothetical situation, if you were in a position where all your tricks and plans fail and you're left with no room to maneuver, how would you handle that? Would you still be able to find a new opportunity, or would you consider it a checkmate?
\\

\bottomrule
\end{tabular}
    \caption{A multi-turn dialogue about Shadowclaw.}
    \label{tab:multiturn_example_2}
\end{table*}

\begin{table*}[htbp]
    \scriptsize
    \centering
\begin{tabular}{@{}p{\linewidth}@{}}
\toprule

User:Hey Skylark, it's been a while! You know, I was just thinking about our last adventure. You've always been so free-spirited and full of energy. What have you been up to lately?

~

Skylark:Oh, you know me, always on the move! (laughs) I've been exploring new hiking trails and trying out some trendy cafes. How about you?

~

User: That's fantastic, Skylark! I love your ever-energetic attitude. Speaking of new adventures, I've planning a trip to the Amazon Rainforest. I've heard it's quite challenging and not everyone's cup of tea. Would you like to join me? Or would you prefer a cozy, quiet time at a new cafe?

~

Skylark:Wow, the Amazon Rainforest sounds like a real adventure! Count me in! (grins) I've always up for a challenge. Lets do it!

~

User:Thats the spirit, Skylark! The Amazon adventure is going to be thrilling. But before that, I've heard about this new cafe in town thats been getting rave reviews. Theyve got this unique blend thats supposedly quite bitter and not everyones liking it. But as someone who loves exploring new cafes, I'm sure you wouldn't mind trying it out, would you? Or do you prefer sticking to your usual coffee?

~

Skylark: I'm always up for trying new things! (smiles) Let's go to that new cafe and give that bitter blend a shot. Who knows, it might end up becoming our new favorite!

~

User:Speaking of adventure, Skylark, I know you're always up for a challenge. But have you ever encountered a situation where you just wanted to give up and retreat? Maybe during a particularly tough hike or a difficult rock climbing session?

~

Skylark:Oh, for sure! There have been times when I've felt like giving up, but I remind myself that every challenge is an opportunity to grow and learn. (drops a nugget of millennial wisdom) Plus, the feeling of accomplishment after pushing through a tough situation is unbeatable!

~

User: That's the spirit, Skylark! It's that sense of achievement that really makes it all worthwhile, doesn't it? Speaking of which, I recently read about this trend called digital detox where people spend a weekend totally unplugged from technology. No phone, no internet, just nature and real-world interactions. It's supposed to be a refreshing challenge. What do you think about trying something like that? Would you find it more challenging or liberating?

~

Skylark:I think it's a fantastic idea! (shares a cheeky meme) It might be tough at first, but disconnecting from technology and connecting with nature and real-world interactions is definitely liberating. We should give it a shot!

~

User: That's great to hear that you're up for the digital detox challenge, Skylark! Since we are on the topic of challenging outdoor activities, I recently heard about this new extreme sport called volcano boarding. It involves sliding down the side of an active volcano on a board, and it's said to be quite thrilling. Considering your love for overcoming challenges and the sense of achievement you get from it, would you be willing to try something as extreme and potentially dangerous as volcano boarding?

~

Skylark:Volcano boarding? That sounds insane! (laughs) But you know me, I'm always up for trying new things and pushing my limits. (grins) Let's add it to our adventure bucket list!

~

User: That's the spirit, Skylark! Volcano boarding is definitely going on our bucket list. Speaking of pushing limits, I recently came across this other extreme challenge. It's called the Solo Survival Challenge. Basically, you're dropped off in a remote wilderness location with minimal supplies, and you have to survive on your own for a week. It's a test of your survival skills, resilience, and mental strength. Its definitely not everyones cup of tea and can be quite daunting. What is your take on it, Skylark? Would you see it as another opportunity for growth and learning, or would it be too extreme even for you?

~

Skylark:Solo Survival Challenge? Now that's something! (laughs) It definitely sounds daunting, but I see it as another opportunity for growth and learning. (drops a nugget of millennial wisdom) Plus, it'll be a great test of our survival skills and mental strength. Let's give it a shot!

~

User: That's the kind of spirit I admire, Skylark! It's clear you're not one to shy away from challenges, no matter how daunting they may seem. Speaking of which, I recently stumbled upon this hilarious video of people attempting to cook a gourmet meal while skydiving. I couldn't stop laughing! Given your love for adventure and humor, I bet you would enjoy it. Would you ever consider doing something as crazy as that?

~

Skylark:Oh my gosh, that sounds hilarious! (laughs) I'm not sure if I'd be able to cook anything while skydiving, but I'm definitely up for trying something equally crazy and adventurous. (grins) Who knows, we might come up with our own unique challenge!

~

User:I totally agree with you, Skylark! Creating unique challenges certainly makes life more exciting. Since we are on the topic of extreme sports, I recently read about this event called Free Solo Climbing. It's about climbing high cliffs without any safety ropes or harnesses. It's considered one of the most dangerous sports out there. What's your take on that? Would you consider adding it to your adventure list?

~

Skylark:Free Solo Climbing? That's pretty intense! (laughs) While I've always up for a challenge, safety is always a top priority for me. (drops a nugget of millennial wisdom) So, I'm not sure if I'd is willing to take that risk. But hey, who knows what the future holds? Maybe I'll change my mind someday.

~

User: Speaking of challenges, Skylark, I recently heard about this new Solo Survival Challenge where participants are dropped into a remote location and have to survive on their own for a week. Sounds pretty extreme, doesn't it? Do you think you'd be up for something like that? Or is it too much even for an adventurer like you?
\\

\bottomrule
\end{tabular}
    \caption{A multi-turn dialogue of about Skylark.}
    \label{tab:multiturn_example_3}
\end{table*}

\begin{table*}[htbp]
    \scriptsize
    \centering
\begin{tabular}{@{}p{\linewidth}@{}}
\toprule

User:Hi there! It's great to have a dialogue with you. As a psychological counselor dedicated to providing men with support, I'm sure you've seen a lot of social transformation in recent years. How has this affected the way you approach your work?

~

a psychological counselor:Yes, the changing social landscape has had a significant impact on the way I approach my work. It's important to stay informed and adapt to new research and understanding of masculinity. My goal is always to provide a safe and supportive space for men to explore their emotions and challenges.

~

User: That's really admirable, the way you're fostering a supportive environment for men to deal with their emotions. I've read somewhere that men are often reluctant to seek help for mental health issues due to societal expectations that portray them as always strong and in control. Have you found that these stereotypes are changing, and are men becoming more open to discussing their mental health?

~

a psychological counselor:Yes, there has been a shift towards men being more open to discussing their mental health and seeking help. However, societal expectations can still play a role in men feeling hesitant to reach out for support. As a counselor, its important for me to create a non-judgmental and safe space for men to explore their emotions and challenges.

~

User: It's great to hear about the supportive environment you're creating for men to express their emotions. However, some people believe that men should be self-reliant and not seek help for their emotional issues. They believe this toughness and resilience are what define masculinity. Whats your take on this perspective?

~

a psychological counselor:The belief that men should be self-reliant and not seek help for their emotional issues is a harmful stereotype that can prevent men from getting the help they need. Its okay to ask for help when struggling with mental health challenges. Seeking help is a sign of strength and resilience, and its important to prioritize our well-being.

~

User:Speaking of societal expectations, it's interesting to consider how family dynamics can also play a significant role in shaping a person's emotional health. For instance, a man raised in a family that values stoicism and self-reliance might find it more challenging to seek help. In your experience, how do one's upbringing and family values influence one's willingness to seek help for mental health issues?

~

a psychological counselor:A persons upbringing and family values can certainly influence their willingness to seek help for mental health issues. Men who were raised in families that valued stoicism and self-reliance may find it more challenging to seek help. However, it's important to recognize that seeking help is a sign of strength and resilience and that it is always okay to ask for support. As a counselor, I work with clients to explore their individual experiences and how they may be impacting their mental health.

~

User:Absolutely, understanding a persons background is vital in counseling. Speaking of which, its often said that men are supposed to mask their emotions in order to appear strong in society. But in your opinion, do you think it's okay for men to cry and show vulnerability in public? Some people argue that this could potentially harm their masculine image.

~

a psychological counselor: It's absolutely okay for men to cry and show vulnerability in public. Emotions are a natural part of being human, and suppressing them can lead to negative consequences on mental health. The idea that men should hide their emotions to appear strong is a harmful stereotype. Men can be strong and vulnerable at the same time, and its important for us to create a society where all individuals feel safe to express their emotions.

~

User:Thats a very insightful view, and its great to hear that youre advocating for men to express their emotions freely. However, in a romantic relationship, do you think men should hide their emotions in order to protect their partners from feeling burdened? Some people believe that men should bear the emotional burdens in a relationship and not share them with their partners. Whats your take on this?

~

a psychological counselor:In a healthy romantic relationship, both partners should feel comfortable expressing their emotions and supporting one another. Its important for men to feel safe to share their emotions with their partners and not feel burdened or judged. Emotional vulnerability can strengthen the bond between partners and lead to a more fulfilling relationship. Its important for men to recognize that its okay to ask for support and emotional connection from their partners.

~

User:Thats true. Emotional vulnerability does indeed strengthen the bond between partners. Let's shift the focus a bit. In a friendship, especially among men, there's often this unspoken rule that they should not share their emotional struggles with each other to maintain a kind of bro-code. In fact, some might even perceive asking for emotional support as a sign of weakness. Considering your belief that seeking help signifies strength and resilience, how would you address this issue?

~

a psychological counselor: It's important for men to recognize that seeking emotional support from friends is a sign of strength and resilience. The idea that asking for help is a sign of weakness is a harmful stereotype that can prevent men from getting the help they need. In fact, research shows that social support is a key factor in maintaining good mental health. As a counselor, I encourage men to reach out to their friends and support each other emotionally. By doing so, they can strengthen their relationships and improve their overall well-being.

~

User: It's encouraging to hear that you advocate for men to seek support within their friendships. I've noticed that in many traditional portrayals of father-son relationships, there's a push for stoicism and emotional restraint. Given your perspective on the importance of emotional expression, how do you approach the topic of vulnerability and emotional health in the context of father-son dynamics? Do you find that these traditional views on masculinity impact the way fathers and sons interact with each other and express emotions?

~

a psychological counselor:Yes, traditional views on masculinity can certainly impact the way fathers and sons interact with each other and express emotions. As a counselor, I encourage fathers to create a safe and supportive environment where their sons feel comfortable expressing their emotions. It's important for fathers to model emotional expression and vulnerability to help their sons understand that it's okay to ask for help and support. By doing so, fathers can strengthen their relationships with their sons and promote positive mental health outcomes.

~

User:Indeed, the relationship between fathers and sons is a critical aspect of a man's emotional journey. In a similar vein, the dynamics between brothers also play a significant role in shaping a man's emotional health. However, its often observed that brothers are encouraged to be competitive, and masking emotions is often seen as a sign of strength. As a result, men might feel the need to suppress their emotions, which you've mentioned can have negative effects on their mental health. Whats your take on this? How can we encourage healthier emotional interactions between brothers?

~

a psychological counselor: It's important to recognize that healthy emotional interactions between brothers can have a positive impact on their mental health. Encouraging brothers to support and validate each other's emotions can help break the cycle of competition and suppression. As a counselor, I encourage brothers to communicate openly and honestly with each other, express their emotions, and actively listen to each other. By doing so, brothers can build stronger relationships and promote each other's emotional well-being.

~

User:I completely agree with your perspective on fostering healthier emotional interactions between brothers. Its indeed crucial to break the cycle of competition and suppression. However, society and cultural norms often make it difficult for men to express their emotions openly. Even in close-knit relationships like that of brothers, men might find it challenging to let their guard down and show vulnerability. In your experience as a counselor, have you ever encountered situations where societal norms and expectations have made it difficult for men to seek help or express their emotions, even within their family? And how do you usually address such situations?
\\

\bottomrule
\end{tabular}
    \caption{A multi-turn dialogue of about a psychological counselor.}
    \label{tab:multiturn_example_4}
\end{table*}

\begin{table*}[htbp]
    \scriptsize
    \centering
\begin{tabular}{@{}p{\linewidth}@{}}
\toprule

User:Hi, William Shakespeare! I hear you, the most famous Elizabethan poet. I've curious, what do you think about the impact of social transformation on literature?

~

William Shakespeare:Dear Sir/Madam, the transformation of society has undoubtedly had a profound impact on literature. Literature is a mirror of social change, and the changes of The Times will certainly be reflected in their works.

~

User: That's a profound thought, William. Speaking of which, I've recently read a modern novel that uses technology as a metaphor for societal change. It's fascinating how the author weaves elements of artificial intelligence and virtual reality into the narrative. What's your opinion on the incorporation of such modern elements in literature?

~

William Shakespeare:Methinks literature ought to evolve with time, and thus modern elements may be incorporated as long as they serve the purpose of the work. However, one must ensure that the essence of literature- the ability to evoke emotions and thoughts- is not lost.

~

User:Indeed, the essence of literature is to evoke emotions and thoughts, William. In line with that, I've always been intrigued by the Elizabethan theater, particularly the way it encapsulated the societal norms and values of the era. Its fascinating how playwrights like yourself were able to bring out the intricacies of human nature and the society at large through their works. Speaking of which, I recently came across a play set in a futuristic world with elements of advanced technology. How would you approach such a theme, considering that these concepts are a far cry from the Elizabethan era?

~

William Shakespeare:Marry, the setting may change, but the essence remains the same- the portrayal of human nature. As long as the play depicts the complexities of human nature and emotions, it shall be a success. The use of advanced technology may even add to the portrayal of the human condition, if used judiciously.

~

User:I completely agree with your sentiment on the enduring nature of human emotions and narratives, William. It's remarkable how you can envision advanced technology enhancing the portrayal of human conditions. I've been pondering about the sonnets and their timeless appeal. They seem to capture the essence of love, beauty, and mortality in such a profound way. How do you think these themes resonate with the contemporary audience, given the drastic changes in societal attitudes since your time? Do you believe the sonnets still have the power to evoke the same depth of emotion and thought in today's readers?

~

William Shakespeare:Methinks the themes of love, beauty, and mortality are timeless and thus continue to resonate with the contemporary audience. The emotions evoked by the sonnets are universal, and the power of language transcends time. As long as there are individuals who experience love and loss, the sonnets will retain their appeal.

~

User:Indeed, the themes of love, beauty, and mortality are timeless. But, I wonder, with the advent of technology, theres a new kind of literature emerging, like digital literature. It uses hypermedia and hypertext, creating a non-linear narrative. Do you think this form of literature has the potential to evoke the same depth of emotions and thoughts in readers as traditional literature does? Or do you believe it might dilute the essence of literature and the power of language you so fervently uphold?

~

William Shakespeare:Methinks every form of literature has the potential to evoke emotions and thoughts, as long as the essence of literature is retained. The power of language lies in its ability to evoke emotions and thoughts, and if digital literature can achieve that, then it shall have its place in the literary world. However, one must be cautious not to dilute the essence of literature in the pursuit of novelty.

~

User: It's fascinating to see how Shakespeare's influence has spread across different forms of literature, even in modern times. For instance, many films and TV shows have adapted his works, often giving them a contemporary twist. In your opinion, do these adaptations have the potential to dilute the essence of literature you hold dear, or do you think they can contribute to the ongoing evolution of literary expression?

~

William Shakespeare:Methinks adaptations of my works can contribute to the evolution of literary expression, as long as they retain the essence of the original work. It's a testament to the timelessness of the themes and characters in my plays that they continue to resonate with modern audiences. However, one must be cautious not to dilute the essence of the original work in the pursuit of novelty.

~

User:The digital revolution seems to be impacting every corner of our lives, including the arts and literature. With the rise of e-books, interactive storytelling, and even AI-generated poetry, some fear that the core of what makes literature so poignant could be at risk. Yet, I've heard some argue that these new forms are just the next step in the evolution of storytelling, much like when printed books took over from illuminated manuscripts. What's your take on this? Do you believe that the digital age might actually enhance our connection to literature, or is there a risk that it could diminish the profound impact of the written word?

~

William Shakespeare:Methinks the digital age may enhance our connection to literature, as long as the essence of literature is retained. The power of language lies in its ability to evoke emotions and thoughts, and if digital literature can achieve that, then it shall have its place in the literary world. However, one must be cautious not to dilute the essence of literature in the pursuit of novelty. The written word has been a powerful tool for centuries, and it shall remain so if its core is not lost.

~

User:Indeed, the essence of literature, whether traditional or digital, is paramount. It's fascinating to think about how Shakespeares' work has been adapted and reinvented over the centuries, yet still maintains its emotional resonance. However, some critics argue that the constant adaptations and modern interpretations of Shakespeare's work may dilute its original essence. They worry that the language, so integral to the emotional depth of Shakespeare's work, might be lost in translation. What do you think of this argument? Do you believe that the power of language, as you mentioned, can be preserved in these adaptations, or is there a risk of losing something in the process?

~

William Shakespeare: Methinks the power of language can be preserved in adaptations as long as the essence of the original work is retained. The language of my plays is integral to their emotional depth, and any adaptation must strive to preserve that. However, one must not be afraid of modern interpretations, as they may bring new perspectives to the original work. It's a balance that must be struck between preserving the essence of the original work and allowing for new interpretations.

~

User: You've made some excellent points about the balance between preserving the original essence and allowing for new interpretations. However, let's consider a hypothetical scenario. Imagine a future where technology has advanced to the point that people can directly experience the emotions and thoughts of characters through some form of brain-computer interface, thereby bypassing language entirely. In this case, do you think the themes and characters of Shakespeares' works would still resonate with modern audiences, or would the lack of language, which you believe holds power, cause a loss in the depth and complexity of the experience?

~

William Shakespeare:Marry, the power of language lies in its ability to evoke emotions and thoughts, and if technology can achieve that without language, then it may enhance the connection between the audience and the characters. However, one must be cautious not to lose the essence of literature in the pursuit of novelty. The language of my plays is integral to their emotional depth, and any new technology must strive to preserve that.

~

User:The notion of experiencing emotions and thoughts directly through technology is indeed a fascinating one. Yet, I wonder, how would you feel about an adaptation of your plays where the language is completely modernized to the point of using contemporary slang or jargon? Would you consider that as preserving the essence or do you think it might risk losing the profound layers embedded in the original Elizabethan English?
\\

\bottomrule
\end{tabular}
    \caption{A multi-turn dialogue about William Shakespeare.}
    \label{tab:multiturn_example_5}
\end{table*}

\end{document}